%% file: main.tex
\documentclass[11pt]{article}

\usepackage[margin=1in]{geometry}
\usepackage{amsmath,amssymb,graphicx}
\usepackage{mathtools}

\usepackage{multirow}
\usepackage{booktabs}
\usepackage{longtable}

\usepackage{lscape}
\usepackage{float}
\usepackage{subcaption}

\usepackage{tikz}
\usetikzlibrary{arrows.meta, positioning, shapes.geometric, fit, backgrounds, calc}

\usepackage[most]{tcolorbox}

\newtcolorbox{systempromptbox}{
  colback=blue!3, colframe=blue!25, colbacktitle=blue!12,
  coltitle=black, fonttitle=\small\bfseries,
  title=System Prompt, rounded corners, boxrule=0.4pt,
  left=4pt, right=4pt, top=2pt, bottom=2pt, fontupper=\small\ttfamily
}
\newtcolorbox{envbox}[1][Environment]{
  colback=orange!3, colframe=orange!25, colbacktitle=orange!12,
  coltitle=black, fonttitle=\small\bfseries,
  title=#1, rounded corners, boxrule=0.4pt,
  left=4pt, right=4pt, top=2pt, bottom=2pt, fontupper=\small\ttfamily
}
\newtcolorbox{agentbox}[1][Agent]{
  colback=gray!4, colframe=gray!30, colbacktitle=gray!12,
  coltitle=black, fonttitle=\small\bfseries,
  title=#1, rounded corners, boxrule=0.4pt,
  left=4pt, right=4pt, top=2pt, bottom=2pt, fontupper=\small\ttfamily
}
\newtcolorbox{oraclebox}[1][Rationality Oracle]{
  colback=green!3, colframe=green!25!black!15, colbacktitle=green!10,
  coltitle=black, fonttitle=\small\bfseries,
  title=#1, rounded corners, boxrule=0.4pt,
  left=4pt, right=4pt, top=2pt, bottom=2pt, fontupper=\small\ttfamily
}

\usepackage[ruled,vlined,linesnumbered]{algorithm2e}
\SetKwInput{KwRequire}{Require}
\SetKwInput{KwEnsure}{Ensure}

\usepackage{xcolor}

\usepackage{hyperref}
\hypersetup{colorlinks=true,linkcolor=blue,citecolor=blue,urlcolor=blue}

\usepackage[round]{natbib}

\input{commands}

\title{OptiRepair: Closed-Loop Diagnosis and Repair of\\Supply Chain Optimization Models with LLM Agents}
\author{
Ruicheng Ao\thanks{Massachusetts Institute of Technology. Email: \texttt{aorc@mit.edu}.} \and
David Simchi-Levi\thanks{Massachusetts Institute of Technology. Email: \texttt{dslevi@mit.edu}.} \and
Xinshang Wang\thanks{Alibaba Group. Email: \texttt{xinshang.w@alibaba-inc.com}.}
}
\date{}

\begin{document}

\maketitle

\begin{abstract}
\input{sections/abstract}
\end{abstract}

\noindent\textbf{Keywords:} supply chain optimization; large language models; model diagnosis and repair; operational rationality; reinforcement learning

\input{sections/introduction}
\input{sections/related}
\input{sections/framework}
\input{sections/training}
\input{sections/experiments}
\input{sections/conclusion}

\bibliographystyle{plainnat}
\bibliography{references}

\clearpage
\appendix

\input{supplement/sections/appendix_a_lp}
\input{supplement/sections/appendix_b_errors}
\input{supplement/sections/appendix_c_oracle}
\input{supplement/sections/appendix_k_prompts}
\input{supplement/sections/appendix_e_full_results}
\input{supplement/sections/appendix_f_guided}
\input{supplement/sections/appendix_i_dataset}

\end{document}

%% file: commands.tex
\newcommand{\Scal}{\mathcal{S}}
\newcommand{\Acal}{\mathcal{A}}

\newcommand{\RR}{\text{RR}}
\newcommand{\RRk}[1]{\text{RR@}{#1}}
\newcommand{\RRR}{\text{RRR}}
\newcommand{\DA}{\text{DA}}

\newcommand{\PPtwo}{\text{P2Pass}}
\newcommand{\IIS}{\text{IIS}}
\newcommand{\BWR}{\text{BWR}}

\newcommand{\OPTIMAL}{\textsc{Optimal}}
\newcommand{\INFEASIBLE}{\textsc{Infeasible}}

\newcommand{\GRPO}{\textsc{GRPO}}
\newcommand{\SFT}{\textsc{SFT}}

\newcommand{\STaR}{\textsc{STaR}}

\newcommand{\OptiSTaR}{\textsc{OptiSTaR}}

\newcommand{\OptiRepair}{\textsc{OptiRepair}}

\newcommand{\ORSC}{\textsc{OptiRepair-SC}}

\newcommand{\Qwen}{\textsc{Qwen3-8B}}
\newcommand{\Llama}{\textsc{Llama-3.1-8B}}

\newcommand{\ME}[1]{\text{ME-#1}}

\newcommand{\st}{\text{s.t.}}
\newcommand{\minimize}{\text{minimize}}

\newcommand{\figcolwidth}{0.75\columnwidth}

\newcommand{\circmark}[1]{\raisebox{.5pt}{\textcircled{\raisebox{-.9pt}{\scriptsize #1}}}}

\newcommand{\figref}[1]{Figure~\ref{fig:#1}}
\newcommand{\tabref}[1]{Table~\ref{tab:#1}}
\newcommand{\secref}[1]{Section~\ref{sec:#1}}

\newcommand{\todo}[1]{}

%% file: sections/abstract.tex
Supply chain optimization models frequently become infeasible because of modeling errors.
Diagnosis and repair require scarce OR expertise: analysts must interpret solver diagnostics, trace root causes across echelons, and fix formulations without sacrificing operational soundness.
Whether AI agents can perform this task remains untested.
\OptiRepair{} decomposes this task into two phases: a domain-agnostic feasibility phase that iteratively repairs any LP using \IIS{}-guided diagnosis, and a domain-specific validation phase that enforces five rationality checks grounded in inventory theory.
We test 22~API models from seven families on 976 multi-echelon supply chain problems and train two 8B-parameter models with self-taught reasoning and solver-verified rewards.
The trained models reach 81.7\% Rational Recovery Rate (\RRR{})---the fraction of problems resolved to both feasibility \emph{and} operational rationality---versus 42.2\% for the best API model and 21.3\% on average.
The gap concentrates in Phase~1 repair, where API models average 27.6\% recovery rate versus 97.2\% for trained models.
Two gaps separate current AI from reliable model repair: solver interaction, as API models restore only 27.6\% of infeasible formulations; and operational rationale, as roughly one in four feasible repairs violate supply chain theory.
Each gap requires a different intervention---targeted training closes the solver interaction gap, while explicit specification as solver-verifiable checks closes the rationality gap.
For organizations adopting AI in operational planning, formalizing what ``rational'' means in their context is the higher-return investment.

%% file: sections/introduction.tex
\section{Introduction}
\label{sec:intro}

Consider a supply chain analyst building a multi-echelon inventory optimization model in Gurobi.
The model spans five echelons and 24 planning periods, with over 500 constraints encoding demand propagation, inventory balance, lead-time coupling, and production capacity.
After submitting the formulation, the solver returns \INFEASIBLE{}.
The analyst must determine \emph{which} constraints conflict, understand \emph{why} (a misspecified lead time? an incorrect capacity limit?), and repair the formulation so that the solution is not merely feasible but operationally sound.
This iterative diagnosis-and-repair loop---\IIS{} analysis, interpretation, model modification, re-solving, and operational validation---consumes hours of expert time and recurs with every model revision.

\IIS{} computation \citep{chinneck2008} identifies \emph{which} constraints conflict but not \emph{why}: an \IIS{} containing an inventory balance equation and a capacity constraint at echelon~3 reveals a mathematical conflict, yet the analyst must still determine whether capacity was understated, demand was overstated, or lead time was misspecified---each calling for a different repair.
Even a correct diagnosis does not guarantee a good repair: relaxing a capacity constraint may restore feasibility but let echelon~3 order ten times its physical throughput---a mathematically valid yet operationally meaningless fix.
Unlike general software debugging, where unit tests sample program behavior, diagnosing and repairing supply chain models demands both mathematical reasoning (interpreting \IIS{} output) and operational reasoning (judging whether the repaired solution makes physical sense).

A feasible, optimal solution is necessary but not sufficient.
A repaired supply chain model must also satisfy what we call \emph{operational rationality}: its solution should exhibit properties consistent with supply chain theory---base-stock structure \citep{clarkscarf1960}, bounded bullwhip amplification \citep{leebullwhip1997, chenbullwhip2000}, inventory allocation, cost consistency \citep{zipkin2000}, and order smoothing.
These five checks form a Rationality Oracle that filters out ``technically correct but operationally absurd'' repairs---a type of error that pure feasibility metrics miss entirely.

LLMs now address operations management tasks \citep{daiswaminathan2025, chen2025manager, aimbench2025}; benchmarks such as NL4Opt \citep{nl4opt2022} and OptiBench \citep{optibench2024} evaluate one-shot formulation, while PILOT-Bench \citep{pilotbench2026} adds multi-step tool interaction but targets formulation, not repair of faulty models.
The \emph{iterative diagnosis-and-repair process} that consumes analyst time in practice remains unevaluated.
LLMs exhibit a 64.5\% self-correction blind spot rate \citep{selfcorrectionbench2025}. Existing correction benchmarks either exclude OR entirely \citep{correctbench2025} or rely on unit tests rather than deterministic solver feedback \citep{swebench2024}.
Can LLMs diagnose and repair supply chain optimization models effectively when given structured solver feedback?

\paragraph{A closed-loop framework linking optimization feasibility to operational decisions.}
\OptiRepair{} splits optimization model diagnosis and repair into two independently deployable phases.
Phase~1 iterates on solver \IIS{} feedback to diagnose and repair constraint conflicts (up to 20 turns); it applies to any LP, not just supply chains.
Phase~2 applies a domain oracle that checks whether the solution is operationally sound; we implement it for supply chains with five theory-grounded rationality checks (\secref{phase2}).
Each phase is independently trainable, but combined they connect \emph{optimization} (``is the model solvable?'') to \emph{operations} (``does the solution make sense?'').
We evaluate the full pipeline on \ORSC{}, 976 supply chain problems spanning 10 error types (\ME{1}--\ME{10}), and benchmark agentic performance against 22 frontier API models (\tabref{positioning}).

\paragraph{A domain-specialized training pipeline that outperforms frontier API models.}
Using iterative self-taught reasoning (\STaR{}) with solver-verified rewards, we train two 8B-parameter models---one for feasibility repair (Phase~1) and one for rationality repair (Phase~2).
The combined pipeline reaches 81.7\% Rational Recovery Rate ($\RRR{}$)---the fraction of problems where the agent restores solver feasibility \emph{and} produces an operationally rational solution.
The best API model (GPT-5.2, tied with Gemini~2.5~Pro) reaches 42.2\%, a gap of 39.5 percentage points.
The gap holds across two base architectures: \Qwen{} at 81.7\% and \Llama{} at 80.3\%, differing by only 1.4 percentage points.

\paragraph{Empirical evidence that domain specialization, not model scale, drives operational value.}
Across 22 API models spanning 7 families, two gaps persist: solver interaction (API average 27.6\% recovery rate versus 97.2\% for trained models) and operational rationale (roughly one in four feasible repairs violate supply chain theory); prompt engineering cannot reliably close either (one model improves by 7.4\,pp, another degrades by 24.1\,pp).
Domain-specialized 8B models outperform all 22 frontier API models; codifying operational rationale as solver-verifiable checks improved results more than scaling model parameters.

\vspace{0.5em}

\tabref{preview} summarizes the central result.
\secref{related} reviews related work; \secref{framework} defines the evaluation framework; \secref{training} details the training pipeline; \secref{experiments} reports agentic performance; and \secref{conclusion} discusses implications and limitations.

\begin{table}[t]
\caption{Central result on \ORSC{} (284-problem test set). $\RRR{}$: fraction of problems where the model restores feasibility \emph{and} produces an operationally rational solution.}
\label{tab:preview}
\centering
\small
\begin{tabular}{@{}lc@{}}
\toprule
Configuration & $\RRR{}$ \\
\midrule
Trained 8B pipeline (best) & 81.7\% \\
Best API (GPT-5.2 / Gemini 2.5 Pro) & 42.2\% \\
Average API (22 models, 7 families) & 21.3\% \\
\bottomrule
\end{tabular}
\end{table}

%% file: sections/related.tex
\section{Related Work}
\label{sec:related}

\OptiRepair{} draws on three research streams: AI and LLMs in operations management, LLM-assisted optimization, and supply chain theory together with verifiable-reward training methods.

\subsection{AI and LLMs in Operations Management}
\label{sec:related_ai_om}

\citet{daiswaminathan2025} organize AI in OM around three pillars: AI for OM (using AI to improve operations), OM for AI (applying operational principles to AI system design, e.g., \citealp{ao2025llminference}), and Human-AI interaction (how humans and AI collaborate on operational tasks).
\OptiRepair{} belongs to the first pillar---training AI agents for an operational task---and advances the third by identifying where AI succeeds (Phase~2 rationality repair, 76.3\% average pass rate) and where it fails (Phase~1 repair, 27.6\% average recovery rate).

Behavioral operations research identifies systematic biases in human operational decisions.
\citet{schweitzer2000decision} established that newsvendor ordering exhibits a pull-to-center bias, and \citet{boltonkatok2008} confirmed that experience reduces but does not eliminate these biases.
Demand amplification studies \citep{sterman1989} and behavioral operations surveys \citep{donohue2020} reinforce the case for automated diagnosis assistance.
Several studies now test whether LLMs replicate these biases.
\citet{chen2025manager} reported that ChatGPT exhibits demand-chasing and pull-to-center biases in newsvendor decisions, replicating patterns \citet{schweitzer2000decision} and \citet{kremer2010demand} observed in human subjects.
AIM-Bench \citep{aimbench2025} tested agentic LLMs as inventory managers in multi-period settings, confirming persistent pull-to-center behavior.
\citet{zhang2025predicting} evaluated LLMs as human-behavior simulators in OM experiments, finding that they predict treatment effects but miss distributional properties of actual decisions.
In concurrent work, \citet{baek2026inventory} benchmarked LLM agents on over 1{,}000 inventory scenarios, finding that OR-augmented LLM policies and human-AI teams each outperform their components in isolation; \citet{simchilevi2026llm} showed that LLMs can serve as interactive interfaces to supply chain solvers, explaining recommendations and adjusting parameters through dialogue; and \citet{long2025beergame} deployed LLM agents on the beer distribution game, observing severe bullwhip effects in decentralized settings.
These studies confirm that LLMs and optimization tools are complementary---a premise \OptiRepair{} applies through solver-in-the-loop repair.
\citet{gijsbrechts2022} applied deep RL to inventory management across dual sourcing, lost sales, and multi-echelon settings, finding that RL policies match or exceed heuristics but require careful reward design.
\citet{davismsom2024} argued for integrating machine learning and behavioral science, since neither approach alone captures how humans make operational decisions.
\citet{disorbo2025} measured how warnings and endorsements shape human-AI collaboration in demand forecasting.
More broadly, integrating ML-generated signals into sequential decisions under imperfect feedback---whether LLM judges replacing human evaluators, prediction-powered optimization, or surrogate rewards for bandits---is an active area \citep{ao2026bai, ao2026ppisvrg, ji2025mabsurrogate}; our Rationality Oracle takes a similar approach, replacing expert review with automated solver-grounded checks.
\OptiRepair{} differs from these lines of work in two ways: it evaluates AI agents on a \emph{structured repair task} with deterministic solver verification rather than prediction or quantity decisions, and it treats \emph{operational rationality} as a first-class metric separate from task completion.

\subsection{LLM-Assisted Optimization}
\label{sec:related_llm_opt}

Multiple benchmarks now evaluate LLMs on optimization tasks.
Most focus on one-shot formulation: NL4Opt \citep{nl4opt2022}, OptiBench \citep{optibench2024}, ORLM \citep{orlm2025}, MAMO \citep{mamo2024}, and others \citep{chainofexperts2024, optimus2024}.
Several testbeds add multi-step interaction: PILOT-Bench \citep{pilotbench2026} evaluates LLM robustness under probabilistic tool failures and variable instruction quality, DP-Bench \citep{dpbench2025} applies solver verification to dynamic programming formulation, and LEAN-LLM-OPT \citep{leanlllmopt2026} coordinates multiple agents to auto-formulate large-scale problems via structured workflow; all target formulation accuracy rather than repairing faulty models.
\citet{selfcorrectionbench2025} measured a 64.5\% self-correction blind spot rate in LLMs; CorrectBench \citep{correctbench2025} benchmarks correction across reasoning domains but excludes operations research.
SWE-bench \citep{swebench2024} benchmarks iterative code debugging, but unit tests only \emph{sample} program behavior, whereas a solver delivers \emph{complete}, deterministic feedback.
\OptiRepair{} addresses this gap: it evaluates iterative \emph{diagnosis and repair} of faulty models with solver feedback, extending the correctness criterion beyond feasibility to domain-specific operational rationality.
\citet{ao2026solverintheloop} introduce a solver-in-the-loop framework that benchmarks diagnostic accuracy of off-the-shelf LLMs on optimization debugging without closing the training loop.
We measure end-to-end agentic performance---whether an agent can repair a formulation \emph{and} validate the resulting solution---rather than diagnostic accuracy alone; apply the framework to multi-echelon supply chain models with theory-grounded rationality checks; and train specialized 8B-parameter agents through iterative \STaR{} with solver-verified rewards.

\input{tables/benchmark_positioning}

\tabref{positioning} contrasts \OptiRepair{} with existing benchmarks on oracle type, multi-step interaction, self-correction, and domain-specific rationality checks.

\subsection{Supply Chain Theory and Training Methods}
\label{sec:related_foundations}

Our error classification and rationality checks build on classical supply chain theory.
\citet{clarkscarf1960} proved that echelon base-stock policies optimize serial multi-echelon systems.
\citet{leebullwhip1997} traced order variance amplification to information distortion, and \citet{chenbullwhip2000} quantified the resulting bullwhip effect.
\citet{zipkin2000} consolidated the foundations of inventory management.
Our ten error types and five rationality checks encode these theoretical properties; \secref{taxonomy} gives the detailed mapping.

Our training approach extends reinforcement learning with verifiable rewards (RLVR).
\citet{deepseekr1} showed that \GRPO{} with outcome verification induces reasoning behavior without step-level supervision; \citet{grpo2024} and \citet{tulu3rlvr} confirmed that RLVR generalizes across reasoning domains.
\citet{star2022} introduced \STaR{} for bootstrapped self-improvement.
We adapt these methods to OR model repair, where Gurobi supplies exact, deterministic verification at every step---feasibility status, objective value, and \IIS{} certificates---enabling fully automated reward computation without human annotation.

%% file: tables/benchmark_positioning.tex
\begin{table}[t]
\caption{Positioning against optimization-agent benchmarks. \ORSC{} is the first to evaluate both iterative diagnosis-and-repair and operational rationality in a closed-loop supply chain setting.}
\label{tab:positioning}
\centering
\footnotesize
\begin{tabular}{@{}llcccc@{}}
\toprule
Framework & Task & Oracle & Multi-step & Self-Corr. & Rationality \\
\midrule
NL4Opt \citep{nl4opt2022}               & NL$\to$LP        & ---             & ---        & ---     & --- \\
OptiBench \citep{optibench2024}          & Formulation      & ---             & ---        & ---     & --- \\
MAMO \citep{mamo2024}                    & Complex LP       & ---             & ---        & ---     & --- \\
ORLM \citep{orlm2025}                    & Formulation      & ---             & ---        & ---     & --- \\
LEAN-LLM-OPT \citep{leanlllmopt2026}    & Formulation      & Solver          & \checkmark & ---     & --- \\
AIM-Bench \citep{aimbench2025}           & Inventory        & Closed-form     & ---        & ---     & \checkmark \\
InventoryBench \citep{baek2026inventory} & Inventory        & Closed-form     & \checkmark & ---     & \checkmark \\
PILOT-Bench \citep{pilotbench2026}       & Tool Workflow    & Execution       & \checkmark & ---     & --- \\
SWE-bench \citep{swebench2024}           & Code Debug       & Unit Tests      & \checkmark & Limited & --- \\
CorrectBench \citep{correctbench2025}    & Self-Corr.       & ---             & ---        & General & --- \\
DP-Bench \citep{dpbench2025}             & DP Problems      & Solver          & \checkmark & ---     & --- \\
\midrule
\ORSC{}                                  & SC Repair        & Solver \IIS{} + Oracle & \checkmark & \checkmark & \checkmark \\
\bottomrule
\end{tabular}
\end{table}

%% file: sections/framework.tex
\section{Closed-Loop Diagnosis and Repair Framework}
\label{sec:framework}

\OptiRepair{} decomposes optimization model repair into a \emph{domain-agnostic} diagnosis-and-repair phase and a \emph{domain-specific} validation phase.
Phase~I applies to any LP: given an infeasible model and solver \IIS{} feedback, the agent iteratively diagnoses constraint conflicts and repairs the formulation.
Phase~II invokes a domain oracle that checks whether the repaired solution satisfies operational criteria; for supply chain models, the oracle verifies properties grounded in inventory theory \citep{clarkscarf1960, chenbullwhip2000}.

\subsection{Multi-Echelon LP Formulation}
\label{sec:lp}

Consider a serial supply chain with $N$ echelons indexed $n = 1, \ldots, N$, where echelon~1 is the retailer (most downstream) and echelon~$N$ is the factory (most upstream).
Material flows from echelon~$N$ toward echelon~1; orders flow in the opposite direction.
The planning horizon spans $T$ discrete periods.

Each echelon~$n$ incurs holding cost $h_n$ and backorder cost $b_n$ per unit per period, faces production or order capacity $C_n$, and has replenishment lead time $L_n$ periods.
Customer demand at the retailer in period~$t$ is $d_t$.
Decision variables are order quantity $x_{n,t}$, on-hand inventory $I_{n,t}$, and backorders $B_{n,t}$ at each echelon and period.
The LP minimizes total holding and backorder costs:
\begin{align}
\minimize \quad & \sum_{n=1}^{N} \sum_{t=1}^{T} \bigl( h_n \, I_{n,t} + b_n \, B_{n,t} \bigr) \label{eq:obj} \\[4pt]
\st \quad
& I_{n,t} - B_{n,t} = I_{n,t-1} - B_{n,t-1} + x_{n,t-L_n} - d_{n,t}, \quad \forall\, n,\, t \label{eq:inv_balance} \\
& x_{n,t} \leq C_n, \quad \forall\, n,\, t \label{eq:capacity}
\end{align}
subject also to demand propagation ($d_{n,t} = x_{n-1,t}$ for $n \geq 2$; $d_{1,t} = d_t$) and non-negativity constraints ($x_{n,t},\, I_{n,t},\, B_{n,t} \geq 0$); Appendix~\ref{sec:app_lp} gives the complete formulation with all constraint families.\phantomsection\label{eq:demand_ext}\phantomsection\label{eq:demand_prop}\phantomsection\label{eq:nonneg}
Constraint~\eqref{eq:inv_balance} enforces flow conservation at each echelon, while demand propagation transmits order variability upstream---the bullwhip effect \citep{leebullwhip1997, chenbullwhip2000, simchilevi2008}.

Our generator produces instances with 2--5 echelons, 12--24 periods, and varying cost structures and demand patterns (stationary, step-change, and seasonal).
Typical instances contain 70--400 decision variables and 60--350 constraints.
Gurobi solves every instance to optimality before error injection, confirming a feasible baseline.

\subsection{Error Classification}
\label{sec:taxonomy}

\input{tables/sc_error_taxonomy}

\tabref{sc_error_taxonomy} defines ten error types (\ME{1}--\ME{10}) covering common modeling mistakes in multi-echelon inventory planning \citep{simchilevi2008, zipkin2000}.
Each type targets a specific LP component; we group them into five categories by the operational subsystem they affect.

\emph{Demand and timing errors} (\ME{1}, \ME{2}) produce multi-echelon \IIS{} patterns requiring upstream--downstream dependency tracing.
\emph{Balance and capacity errors} (\ME{3}, \ME{4}): \ME{3} forces an impossible inventory balance state (localized, easy-to-identify \IIS{} signal), while \ME{4} sets capacity far below demand, producing cascading infeasibility across periods.
\emph{Cost and structural errors} (\ME{5}, \ME{6}): \ME{5} distorts holding costs, yielding feasible but operationally irrational solutions; \ME{6} adds bullwhip-inducing constraints that cascade through the echelon structure.
\emph{Coefficient and sign errors} (\ME{7}, \ME{8}) perturb arrival coefficients or invert demand propagation signs, creating multi-constraint \IIS{} patterns.
\emph{Constraint and index errors} (\ME{9}, \ME{10}): \ME{9} adds redundant order constraints; \ME{10} corrupts time indices in demand propagation, producing misleading \IIS{} signals.

A saboteur module generates all errors programmatically: given a valid LP instance, it selects an error type, applies a controlled perturbation, and verifies that (i) the sabotaged model returns \INFEASIBLE{} (or \OPTIMAL{}-but-irrational for \ME{5}), (ii) the \IIS{} is non-empty, and (iii) a ground-truth fix restores both feasibility and rationality.

\subsection{Two-Phase Closed-Loop MDP}
\label{sec:mdp}

\OptiRepair{} uses a two-phase closed-loop architecture (\figref{closed_loop}).
Phase~I is domain-agnostic: the same state space, action space, and transitions apply to any infeasible LP regardless of domain.
Phase~II applies a domain-specific oracle that checks operational correctness.
Each phase operates as a self-contained decision loop with its own reward signal; combined, they connect optimization (``is this model solvable?'') to operations (``does this solution support sound decisions?'').
We model both phases as a deterministic Markov Decision Process $(\Scal, \Acal, T, R, \gamma)$.

\input{figures/fig_closed_loop}

\subsubsection{Phase I: Diagnose and Repair.}
\label{sec:phase1}

The agent receives an infeasible LP and a natural-language description of the intended model.
At each step, the agent observes the state, selects an action, and the environment executes it via Gurobi.

\paragraph{State space.}
A state $s \in \Scal$ is a tuple:
\begin{equation}
s = (\text{problem}_\text{NL},\; \text{code},\; \text{status},\; \text{iis},\; \text{history},\; t),
\label{eq:state}
\end{equation}
where $\text{problem}_\text{NL}$ is the natural-language problem description, $\text{code}$ is the current Gurobi model code, $\text{status} \in \{\INFEASIBLE, \OPTIMAL, \ldots\}$ is the solver return status, $\text{iis}$ lists the constraints and variable bounds in the irreducible infeasible subsystem, $\text{history}$ records previous actions and outcomes, and $t$ counts steps.

\paragraph{Action space.}
Actions $a \in \Acal$ split into three categories: \emph{diagnostic} actions (\texttt{GET\_IIS}, \texttt{CHECK\_SLACK}) that query the solver without modifying the model; \emph{repair} actions (\texttt{RELAX\_CONSTRAINT}, \texttt{DROP\_CONSTRAINT}, \texttt{UPDATE\_OBJ}, \texttt{UPDATE\_BOUNDS}, \texttt{UPDATE\_RHS}) that modify the model; and the \emph{meta} action \texttt{SUBMIT} that terminates the episode.
Each action specifies a target constraint or variable by name and, where applicable, a numerical value.

\paragraph{Transition.}
Transitions are deterministic: given state $s$ and action $a$, the environment applies $a$ to the Gurobi model, re-solves, recomputes the \IIS{} if still infeasible, and returns $s'$.
Phase~I succeeds when the solver returns \OPTIMAL{}.
Exhausting the 20-step budget without restoring feasibility ends the episode as a Phase~I failure.
\figref{phase1_example} illustrates a Phase~I interaction: the agent diagnoses an \ME{4} error (capacity too tight) using \texttt{GET\_IIS} and repairs it in a single step.

\input{figures/fig_phase1_example}

\subsubsection{Phase II: Validate.}
\label{sec:phase2}

Once Phase~I reaches \OPTIMAL{}, the repaired model enters Phase~II.
A replaceable domain oracle encodes expert criteria for operational correctness beyond mathematical feasibility.
For multi-echelon supply chain models, our \emph{Rationality Oracle} enforces five checks grounded in inventory theory \citep{clarkscarf1960, chenbullwhip2000}: (1)~\emph{base-stock structure} (inventory levels approximate echelon base-stock policies), (2)~\emph{bullwhip ratio} (order variance amplification stays within theoretical bounds \citep{dejonckheere2003}), (3)~\emph{inventory allocation} (downstream echelons do not hoard disproportionate inventory), (4)~\emph{cost consistency} (holding cost ordering and objective coefficients match the problem specification), and (5)~\emph{order smoothing} (order quantities avoid extreme oscillation).
Appendix~\ref{sec:app_oracle} gives formal definitions and thresholds.

Not all checks apply to every error type: an error-type-aware mapping excludes checks whose violations reflect saboteur-injection artifacts rather than genuine irrationality (Appendix~\ref{sec:app_oracle} gives the full mapping).
When any applicable check fails, the agent receives natural-language feedback identifying the violated property and re-enters the repair loop for up to three iterations.
Phase~II succeeds when all applicable checks pass; it fails if iterations run out or a repair breaks feasibility.
\figref{phase2_example} shows a Phase~II loop-back: the oracle detects a cost-consistency violation, the agent corrects the objective coefficient, and the oracle confirms all checks pass.

\input{figures/fig_phase2_example}

\subsubsection{Combined reward.}

The episode reward $R = R_{\text{outcome}} + R_{\text{rationality}}$ sums a feasibility term ($+100$ if \OPTIMAL{}, $-50$ otherwise) and a rationality term ($+50$ if all checks pass, $-25$ if \OPTIMAL{} but checks fail, $0$ if \INFEASIBLE{}):
\begin{equation}
R =
\begin{cases}
+150 & \text{\OPTIMAL{} and all rationality checks pass}, \\
+75 & \text{\OPTIMAL{} but one or more checks fail}, \\
-50 & \text{\INFEASIBLE{} (Phase~I failure)}.
\end{cases}
\label{eq:reward}
\end{equation}
The agent therefore prioritizes feasibility restoration while still earning credit for solution quality.

\subsection{Evaluation Metrics}
\label{sec:metrics}

Three complementary metrics measure repair performance.

\paragraph{Recovery Rate (\RR{}).}
The fraction of problems where the agent restores the solver to \OPTIMAL{} status within the step budget:
\begin{equation}
\RR = \frac{|\{i : \text{status}_i = \OPTIMAL\}|}{N}.
\label{eq:rr}
\end{equation}
\RR{} isolates raw repair ability: can the agent find \emph{any} feasible modification, regardless of solution quality?

\paragraph{Rational Recovery Rate (\RRR{}).}
The fraction of problems where the agent achieves \OPTIMAL{} \emph{and} passes all applicable rationality checks:
\begin{equation}
\RRR = \frac{|\{i : \text{status}_i = \OPTIMAL \;\land\; \text{rational}_i = \text{True}\}|}{N}.
\label{eq:rrr}
\end{equation}
\RRR{} is the primary metric for \ORSC{}, encoding both repair skill (Phase~I) and operational understanding (Phase~II) in a single number.

\paragraph{Phase 2 Pass Rate (\PPtwo{}).}
Among problems where the agent achieves \OPTIMAL{}, the fraction that also passes rationality checks: $\PPtwo = \RRR / \RR$.\label{eq:p2pass}
An agent with high \RR{} but low \PPtwo{} restores feasibility yet yields operationally nonsensical solutions.
Decomposing $\RRR = \RR \times \PPtwo$ separates repair skill from operational understanding; \secref{experiments} reports all three metrics.

\subsection{Problem Set Construction}
\label{sec:benchmark}

\input{figures/fig_benchmark_pipeline}

\paragraph{\ORSC{}.}
\ORSC{} (\OptiRepair{} Supply Chain) contains 976 instances across all ten error types (\ME{1}--\ME{10}), with disjoint supply chain configurations between the training and test splits (Appendix~\ref{sec:app_dataset} details the split).
Each problem bundles a natural-language description, Gurobi Python code for the sabotaged LP, the ground-truth fix, and the error type label.
Instances span 2--5 echelons, 12--24 periods, three demand patterns, holding-to-backorder ratios from $1{:}2$ to $1{:}10$, and both capacitated and uncapacitated variants.
We validate every problem: the sabotaged code returns \INFEASIBLE{} (or \OPTIMAL{}-but-irrational for \ME{5}), and the ground-truth fix passes all applicable rationality checks.
We report phase-wise performance on \ORSC{} through \RR{} (Phase~I recovery), \PPtwo{} (Phase~II conditional pass), and \RRR{} (end-to-end success).

%% file: tables/sc_error_taxonomy.tex
\begin{table}[t]
\caption{Supply chain error classification. Ten error types (\ME{1}--\ME{10}) represent common modeling mistakes in multi-echelon inventory planning.}
\label{tab:sc_error_taxonomy}
\centering
\small
\begin{tabular}{@{}llll@{}}
\toprule
Type & Error Name & Operational Meaning & Difficulty \\
\midrule
\ME{1}  & Demand Inflation       & Overstated downstream demand        & Hard \\
\ME{2}  & Lead Time Error        & Misspecified replenishment lead time & Hard \\
\ME{3}  & Balance Violation      & Impossible inventory balance state   & Easy \\
\ME{4}  & Capacity Reduction     & Production capacity below demand     & Hard \\
\ME{5}  & Cost Structure Error   & Non-monotonic holding costs          & Easy \\
\ME{6}  & Bullwhip Amplification & Forced upstream order amplification  & Medium \\
\ME{7}  & Coefficient Perturbation & Distorted arrival coefficient      & Medium \\
\ME{8}  & Sign Error             & Inverted demand propagation sign     & Medium \\
\ME{9}  & Redundant Constraint   & Conflicting minimum-order bound      & Medium \\
\ME{10} & Index Mismatch         & Wrong time index in demand propagation & Hard \\
\bottomrule
\end{tabular}
\end{table}

%% file: figures/fig_closed_loop.tex
\begin{figure}[t]
\centering
\includegraphics[width=0.55\columnwidth]{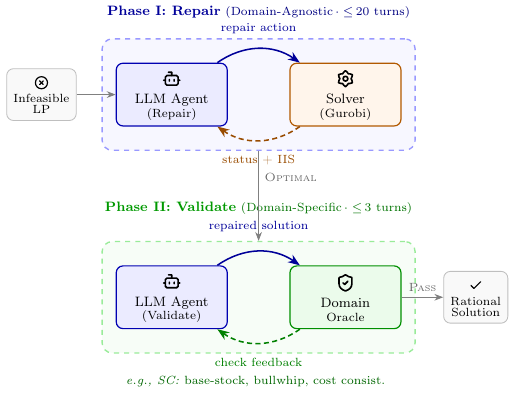}
\caption{Two-phase closed-loop architecture of \OptiRepair{}.
Phase~I (domain-agnostic) iteratively diagnoses and repairs an infeasible LP via solver \IIS{} feedback.
Upon reaching \OPTIMAL{}, Phase~II (domain-specific) validates operational correctness through a replaceable domain oracle---implemented here with five supply chain rationality checks.}
\label{fig:closed_loop}
\end{figure}

%% file: figures/fig_phase1_example.tex
\begin{figure}[!ht]
\centering
\includegraphics[width=0.8\columnwidth]{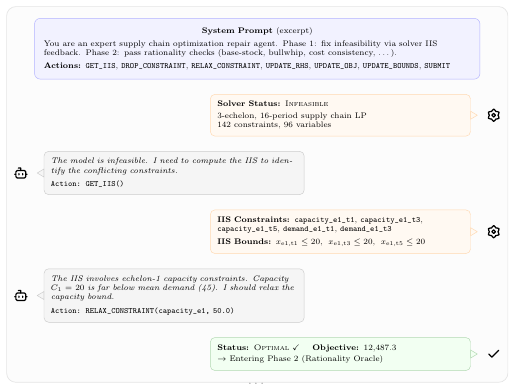}
\caption{Phase~I repair interaction.
The agent receives an infeasible supply chain LP
(Gurobi),
uses \texttt{GET\_IIS} to diagnose conflicting constraints,
and repairs the capacity bound.
When the solver returns \OPTIMAL{}, the episode transitions to Phase~II.
Full system prompt and transcript appear in Appendix~\ref{sec:app_prompts}.}
\label{fig:phase1_example}
\end{figure}

%% file: figures/fig_phase2_example.tex
\begin{figure}[!ht]
\centering
\includegraphics[width=0.75\columnwidth]{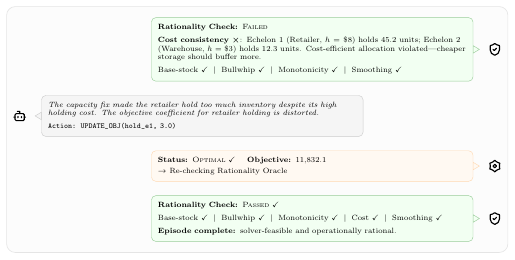}
\caption{Phase~II rationality repair interaction.
The rationality oracle
detects that the Phase~I repair created a cost-consistency violation:
the high-cost retailer holds more inventory than the cheaper warehouse.
The agent corrects the objective coefficient; the solver re-optimizes
and the oracle confirms all five checks pass.}
\label{fig:phase2_example}
\end{figure}

%% file: figures/fig_benchmark_pipeline.tex
\begin{figure}[t]
\centering
\includegraphics[width=\columnwidth]{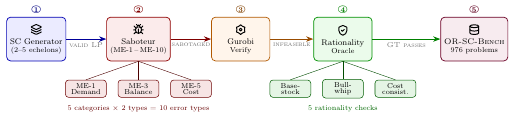}
\caption{\ORSC{} construction pipeline.
\circmark{1}~A multi-echelon supply chain generator produces valid LPs with 2--5 echelons and 12--24 periods.
\circmark{2}~A saboteur injects one of ten error types, grouped into five categories (demand/timing, balance/capacity, cost/structure, coefficient/sign, constraint/index).
\circmark{3}~Gurobi confirms the sabotaged model is infeasible (or \OPTIMAL{}-but-irrational for \ME{5}).
\circmark{4}~The ground-truth fix is verified against five rationality checks.
\circmark{5}~Validated instances form \ORSC{} (976 problems, disjoint train/test splits).}
\label{fig:benchmark_pipeline}
\end{figure}

%% file: sections/training.tex
\section{Training Pipeline}
\label{sec:training}

Two specialized 8B-parameter models drive the \OptiRepair{} closed-loop pipeline---one per phase.
The training procedure, \OptiSTaR{}, cycles through solver-seeded \STaR{} exploration and \GRPO{} refinement at each iteration.

\subsection{Two-Model Architecture}
\label{sec:two_model}

The \OptiRepair{} pipeline uses two independently trained 8B-parameter models: a \emph{P1 model} for Phase~I repair (\INFEASIBLE{} $\to$ solver-feasible) and a \emph{P2 model} for Phase~II rationality repair (solver-feasible $\to$ operationally rational).
Each model solves a self-contained problem with its own reward signal: P1 maximizes solver feasibility recovery using \IIS{} feedback, while P2 maximizes oracle check compliance using domain-specific rationality criteria.
Separating the two objectives isolates feasibility restoration from operational validation.
Phase~I demands LP structure and \IIS{}-based diagnosis under multi-turn solver interaction.
Phase~II demands domain-specific knowledge (base-stock structure, bullwhip bounds, cost consistency) and can be swapped or extended without retraining Phase~I.
A joint model mixes both objectives: it can restore feasibility by deleting cost constraints, producing irrational solutions.

Both models initialize from \Qwen{}-Instruct, selected via a pilot study (+41.9\% $\RRk{5}$ headroom).
At inference, P1 runs for up to 20 turns; once the solver returns \OPTIMAL{}, P2 takes over for up to 3 rationality repair iterations.

\subsection{Iterative \STaR{} for Phase~I Repair}
\label{sec:star}

\input{figures/fig_training_pipeline}

Phase~I training follows an iterative \STaR{} loop \citep{star2022}: beam search exploration \citep{wei2022cot}, supervised distillation, and \GRPO{} refinement (\figref{training_pipeline}).

\paragraph{Stage 1: Beam Search Exploration.}
The current model generates $K=32$ candidate trajectories per problem; the shortest trajectory that restores \OPTIMAL{} status with $\DA \geq 0.5$ survives.

\paragraph{Stage 2: Supervised Distillation.}
\SFT{} on successful beam trajectories with completion-only loss distills search-time gains into the model's greedy policy.

\paragraph{Stage 3: \GRPO{} Refinement.}
Group Relative Policy Optimization \citep{grpo2024, deepseekr1} refines the distilled model. For each problem, the model generates $G=8$ rollout trajectories; rewards normalize within each group to yield advantages. The composite reward balances three objectives:
\begin{equation}
R = 0.5 \cdot R_{\text{outcome}} + 0.3 \cdot R_{\text{diagnosis}} + 0.2 \cdot R_{\text{efficiency}},
\label{eq:composite_reward}
\end{equation}
where $R_{\text{outcome}} = +100$ if the solver returns \OPTIMAL{} and $-50$ otherwise; $R_{\text{diagnosis}} = \DA \times 100$, rewarding overlap between the model's targeted constraints and the ground-truth \IIS{}; and $R_{\text{efficiency}} = -1$ per repair step, penalizing long trajectories. A faithfulness penalty of $-20$ fires when the model targets constraints outside the \IIS{}, discouraging indirect fixes that mask the root cause.

Hyperparameters: KL coefficient $\beta = 0.02$, asymmetric clipping $[\epsilon_{\text{low}}, \epsilon_{\text{high}}] = [0.2, 0.28]$, LoRA rank $r=16$, $\alpha=32$.

The \STaR{} loop ran three iterations, seeded with 696 successful trajectories from GPT-5.2, o4-mini, and DeepSeek-R1 teachers.
Beam trajectories for hard error types (\ME{2}, \ME{4}) are filtered against ground-truth rationality: trajectories whose repairs fail rationality checks are replaced with upsampled ground-truth traces.
Over the full loop, Phase~I $\RRk{5}$ rises from 44.4\% to 75.7\% and P1 $\RR$ reaches 97.2\%.

\GRPO{}'s direct effect on greedy performance is marginal (+0.7pp in $\RRk{5}$). Its value is indirect: refinement raises beam search solve rate from 81.4\% to 98.3\%, feeding higher-quality distillation data into the next \STaR{} iteration.
In multi-turn agentic settings, RL amplifies data quality rather than directly improving the greedy policy.

\subsection{Oracle-Based Phase~II Training}
\label{sec:p2_training}

Phase~II training teaches the model to repair \emph{operationally irrational} solutions that remain mathematically feasible. Three oracle design decisions shape training effectiveness: error-type-aware check activation, artifact-check relaxation, and prefix-matched constraint operations.

Phase~II \GRPO{} grounds its reward in the solver: each rollout trajectory runs in Gurobi and the Rationality Oracle scores it, tying the training signal to operational quality rather than textual similarity.
The reward is binary (pass/fail), using LoRA $\alpha = 0.5$.
Paired with \GRPO{}-trained Phase~I, the final pipeline reaches $\RRR = 81.7\%$ on the held-out test set.
\ME{5} errors (pure Phase~II problems) reach 100\% $\RRR$; the hardest errors, \ME{2} and \ME{4}, require both phases and yield the largest \GRPO{} gains: from 6\% and 27\% to 64.5\% and 61.3\%, respectively (\secref{exp_per_me}).

\subsection{Architecture Generalization}
\label{sec:arch_gen}

Replicating the full pipeline on \Llama{}-Instruct yields $\RRR = 80.3\%$ on the test set---within 1.4 percentage points of \Qwen{}'s 81.7\%, with identical Phase~I recovery rates (97.2\%).
The ablation pattern also holds: replacing trained P1 with Claude Opus~4.5 drops $\RRR$ from 80.3\% to 40.5\%.
Full results appear in \secref{exp_arch}.
Because the two architectures differ in tokenizer, pre-training data, and instruction-tuning methodology, this consistency confirms that the gains stem from the training procedure and oracle specification rather than architecture-specific properties.

%% file: figures/fig_training_pipeline.tex
\begin{figure}[t]
\centering
\includegraphics[width=0.8\columnwidth]{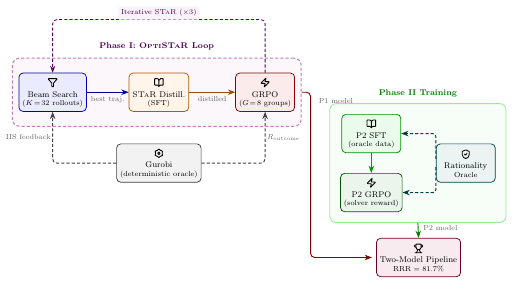}
\caption{\OptiSTaR{} training pipeline.
\emph{Top}: Phase~I iterates through beam search exploration ($K\!=\!32$ candidates), \STaR{} distillation, and \GRPO{} refinement; three iterations raise $\RRk{5}$ from 44.4\% to 75.7\%.
Gurobi provides \IIS{} feedback for beam search and outcome rewards for \GRPO{}.
\emph{Middle right}: Phase~II trains independently on oracle-generated data (SFT) refined by solver-based \GRPO{}.
\emph{Bottom}: The two independently trained 8B models combine into the final pipeline ($\RRR = 81.7\%$).}
\label{fig:training_pipeline}
\end{figure}

%% file: sections/experiments.tex
\section{Agentic Performance}
\label{sec:experiments}

Evaluation focuses on \ORSC{}, a supply chain closed-loop benchmark with two-phase interaction.

\subsection{Experimental Setup}
\label{sec:exp_setup}

\textbf{Benchmark.}
\ORSC{} tests the full two-phase closed-loop pipeline on 284 test problems (from the 976-problem set) across ten supply chain error types (\ME{1}--\ME{10}), with a disjoint training split for domain-specialized models.

\textbf{Models.}
The evaluation covers 24 models: 22 API models from seven families---OpenAI (7), Google (3), Anthropic Claude (4), DeepSeek (2), Qwen (4), Meta (1), and Moonshot (1)---plus two trained 8B-parameter pipelines (\Qwen{} and \Llama{}).

\textbf{Infrastructure.}
All local models run on 2$\times$H100 80GB GPUs, served via SGLang with tensor parallelism (TP=2) and 16-way concurrent evaluation.
API models connect through provider endpoints with rate limiting.
All evaluations use greedy decoding (temperature~$=$~0).
Phase~1 permits up to 20 interaction steps; Phase~2, up to 3 rationality validation iterations.

\subsection{\ORSC{} Results: Supply Chain Closed-Loop}
\label{sec:exp_sc_main}

\input{tables/sc_api_results}
\input{tables/sc_trained_results}

\begin{figure}[t]
\centering
\includegraphics[width=\figcolwidth]{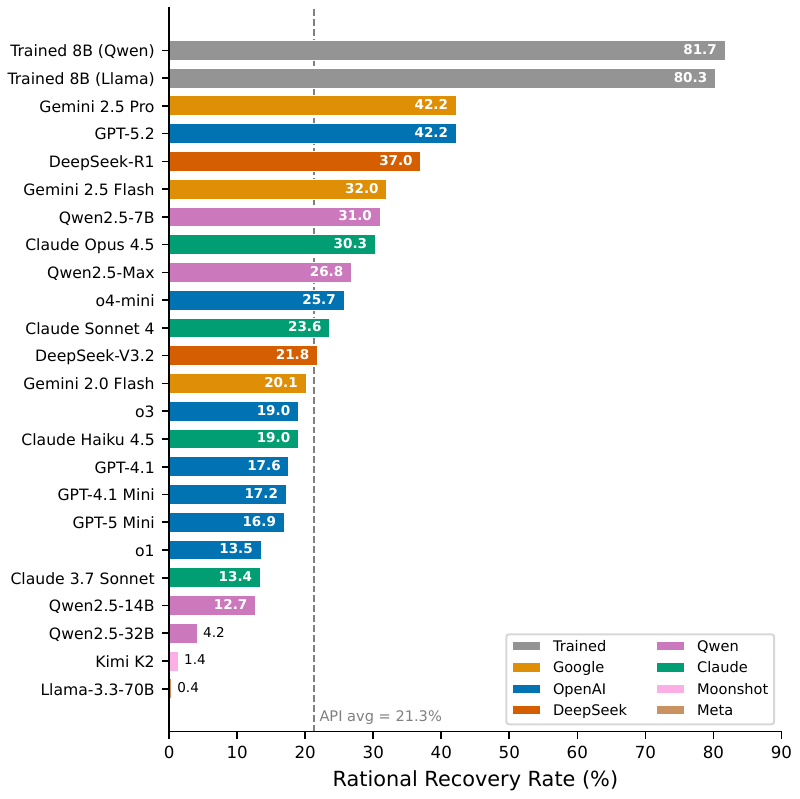}
\caption{Rational Recovery Rate (\RRR{}) for 22 API models and two trained 8B pipelines (\Qwen{} and \Llama{}) on \ORSC{} (284-problem test set). The vertical dashed line marks the API average (21.3\%). Both trained models outperform the best API by $\geq$38 percentage points.}
\label{fig:rrr_bar}
\end{figure}

The trained \Qwen{} pipeline reaches 81.7\% \RRR{} (\tabref{sc_trained_results}), beating the best API models---GPT-5.2 and Gemini~2.5~Pro, both at 42.2\% (\tabref{sc_api_results})---by 39.5 percentage points; the average API model manages only 21.3\%.
Performance varies widely across the 22 API models (\figref{rrr_bar}).
At the bottom, Qwen2.5-32B (4.2\% \RRR{}), Kimi~K2 (1.4\%), and Llama-3.3-70B (0.4\%) fall below 5\% due to Phase~2 format compliance failures.
GPT-5.2, Gemini~2.5~Pro, and DeepSeek-R1-0528 all exceed 37\% \RRR{}, each with Phase~1 recovery above 53\%.
Ranks 4--12 cluster between 19\% and 32\%; model family and parameter count are weak predictors of performance.

\begin{figure}[t]
\centering
\includegraphics[width=0.5\textwidth]{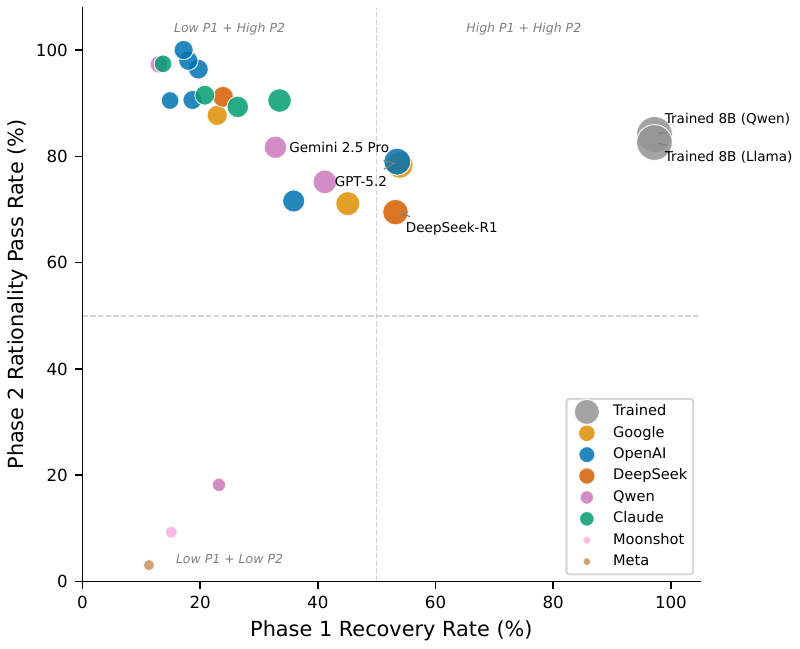}
\vspace{-1em}
\caption{Phase~1 Recovery Rate vs.\ Phase~2 Pass Rate for all 22 API models and two trained 8B pipelines. Bubble size $\propto$ \RRR{}. API models cluster upper-left (high \PPtwo{}, low \RR{}); trained models reach the upper-right corner. The vertical gap between clusters---roughly 70 percentage points in \RR{}---confirms that Phase~1 repair drives the overall performance difference.}
\label{fig:p1_vs_p2}
\end{figure}

The gap concentrates in Phase~1: API models average 27.6\% \RR{} versus 97.2\% for trained models---a 69.6 percentage point gap (\figref{p1_vs_p2}).
This low Phase~1 rate reflects harder repair challenges: longer constraint chains, multiple interacting echelons \citep{simchilevi2008}, and domain-specific variable naming.
Once feasibility is restored, the validation phase poses a lower barrier: the 19 format-compliant API models average 86.7\% \PPtwo{}, with several above 90\%.
The bottleneck is finding a feasible repair, not judging its operational quality.

Model families exhibit distinct repair profiles.
Claude models achieve the highest \PPtwo{} (89--97\%) but the lowest \RR{} among top-tier models (13--34\%); GPT-5.2 and Gemini~2.5~Pro balance both (\RR{} $>$53\%, \PPtwo{} $\approx$79\%).
Reasoning-specialized models (o3: 19.0\%, o1: 13.5\%) do not outperform general models.
DeepSeek-R1-0528 ranks third (37.0\%) with strong \RR{} (53.2\%) but lower \PPtwo{} (69.5\%), illustrating a feasibility--rationality trade-off: aggressive repairs restore solvability at the cost of operational quality.

\subsection{Per-Error-Type Analysis}
\label{sec:exp_per_me}

\tabref{sc_per_me} (Appendix~\ref{sec:app_full_results}) breaks down \RRR{} by error type for the top-6 API models and both trained models.
The ten error types range from universally solvable to near-impossible for API models, with a sharp boundary between errors that produce direct diagnostic signals and those requiring multi-echelon causal reasoning.

All top models score 90--100\% \RRR{} on \ME{5} (cost structure error): the model is already feasible, and the rationality oracle directly identifies the non-monotonic holding costs, making the repair unambiguous.

\ME{2} (lead time error) and \ME{4} (capacity reduction) separate trained from API models.
On \ME{2}, the best API model (GPT-5.2) reaches 41.9\% \RRR{} vs.\ trained \Qwen{} at 64.5\%; on \ME{4}, GPT-5.2 manages 38.7\% vs.\ \Llama{} at 74.2\%.
These errors require tracing how lead-time coupling and capacity constraints interact with inventory balance equations across echelons---domain knowledge that general models lack.

\ME{3} (balance violation) tops the API success ranking (63--77\% for top models) because the \IIS{} directly pinpoints the violated constraint and the repair is localized.
\ME{1} (demand inflation) and \ME{10} (index mismatch) hover near zero for API models (0--7\%), as these errors generate misleading \IIS{} signals that require backward reasoning through the echelon structure.
The pattern is clear: when the \IIS{} points directly to the root cause (\ME{3}, \ME{5}), general models suffice; when diagnosis requires tracing interactions across echelons (\ME{1}, \ME{2}, \ME{4}, \ME{10}), only specialized training closes the gap.

\subsection{Component Ablation}
\label{sec:exp_ablation}

\input{tables/sc_ablation_qwen}

To isolate each phase's contribution, we substitute trained models with API alternatives (\tabref{sc_ablation}).
Replacing the trained Phase~1 model cuts \RRR{} by 22--40pp; replacing only Phase~2 costs 9--13pp.
The trained Phase~1 model reaches 97.2\% \RR{}, exceeding both API alternatives by 31--42pp.
API models perform adequately on Phase~2 because operational reasoning aligns more closely with their general capabilities than multi-step repair does.
End-to-end, GPT-5.2 scores 42.3\% \RRR{} and Opus~4.5 scores 30.3\%---Phase~1 repair drives the gap.
For practitioners building in-house repair tools, Phase~1 capability is the higher-return investment.

\subsection{Architecture Generalization}
\label{sec:exp_arch}

To test architecture dependence, we replicate the full training pipeline on \Llama{} (Llama-3.1-8B-Instruct).
\Llama{} scores 80.3\% \RRR{} versus \Qwen{}'s 81.7\%---a 1.4pp gap---with identical Phase~1 recovery rates (97.2\%).
The ablation patterns replicate across architectures: trained Phase~1 outperforms API Phase~1 by 30--46pp, trained Phase~2 outperforms API Phase~2 by 9--13pp, confirming that the training procedure, not architecture-specific properties, drives effectiveness.

\subsection{Prompt Engineering Limitations}
\label{sec:exp_prompt}

Two guided prompt variants augment the baseline API prompt: an \emph{anti-loop} variant with explicit loop-avoidance instructions, and a \emph{domain-knowledge} variant with supply chain structural hints (base-stock intuition, echelon coupling; full text in Appendix~\ref{sec:app_guided}).
The domain-knowledge prompt raises Claude Opus~4.5 by 7.4 percentage points but degrades GPT-5.2 by 24.1 percentage points; the anti-loop prompt hurts both models.
Prompt engineering alone cannot close the 38--40 percentage point gap to domain-specialized training.

\subsection{Repair Efficiency}
\label{sec:exp_efficiency}

\begin{figure}[t]
\centering
\includegraphics[width=0.5\textwidth]{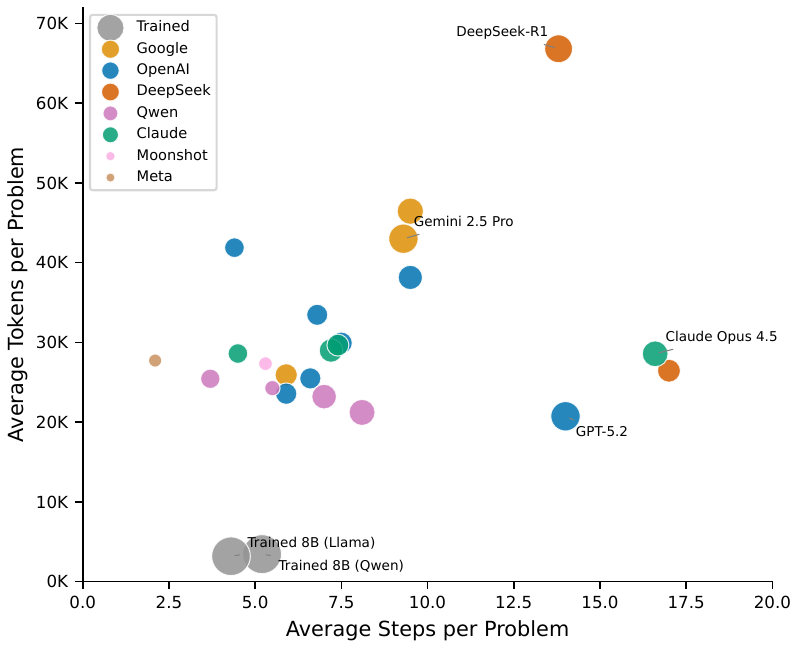}
\vspace{-1em}
\caption{Repair efficiency: average steps vs.\ average tokens per problem for all 24 models. Bubble size $\propto$ \RRR{}. Trained 8B models cluster in the bottom-left corner---fewer steps, fewer tokens, higher \RRR{}---while the highest-\RRR{} API models (GPT-5.2, DeepSeek-R1) consume the most resources.}
\label{fig:efficiency}
\end{figure}

Trained models also repair more efficiently (\tabref{sc_trained_results}, \figref{efficiency}).
The \Qwen{} pipeline averages 5.2 steps and 3,439 tokens per problem; GPT-5.2 requires 14.0 steps and 20,730 tokens, and the 22-model API average is 7.8 steps and 30,078 tokens.
This 2--3$\times$ step reduction and 6--19$\times$ token reduction accompany the 38--40 percentage point \RRR{} gap documented above.
Trained models converge on targeted repairs instead of cycling through equivalent constraint relaxations, repeating similar fixes, or generating verbose diagnostic text---behaviors that consume steps without progress.
Models that solve problems need fewer attempts to do so.

%% file: tables/sc_api_results.tex
\begin{table}[t]
\caption{\ORSC{} results: 22 API models on the supply chain closed-loop pipeline (284 test problems). Models ranked by Rational Recovery Rate (\RRR{}).}
\label{tab:sc_api_results}
\centering
\footnotesize
\setlength{\tabcolsep}{2.5pt}
\begin{tabular}{@{}rlrrrr@{\hskip 8pt}|@{\hskip 8pt}rlrrrr@{}}
\toprule
\# & Model & \RR & \RRR & \PPtwo & Steps &
\# & Model & \RR & \RRR & \PPtwo & Steps \\
\midrule
1  & GPT-5.2           & 53.5 & 42.2 & 79.0 & 14.0 &
12 & Claude Haiku 4.5  & 20.8 & 19.0 & 91.5 & 7.4 \\
2  & Gemini 2.5 Pro    & 53.9 & 42.2 & 78.4 & 9.3 &
13 & GPT-o3            & 19.7 & 19.0 & 96.4 & 7.5 \\
3  & DeepSeek-R1-0528  & 53.2 & 37.0 & 69.5 & 13.8 &
14 & GPT-4.1           & 18.0 & 17.6 & 98.0 & 5.9 \\
4  & Gemini 2.5 Flash  & 45.1 & 32.0 & 71.1 & 9.5 &
15 & GPT-4.1 Mini      & 17.2 & 17.2 & 100.0 & 6.6 \\
5  & Qwen2.5-7B        & 41.2 & 31.0 & 75.2 & 8.1 &
16 & GPT-5 Mini        & 18.7 & 16.9 & 90.6 & 6.8 \\
6  & Claude Opus 4.5   & 33.5 & 30.3 & 90.5 & 16.6 &
17 & GPT-o1            & 14.9 & 13.5 & 90.5 & 4.4 \\
7  & Qwen2.5-Max       & 32.8 & 26.8 & 81.7 & 7.0 &
18 & Claude 3.7 Sonnet & 13.7 & 13.4 & 97.4 & 4.5 \\
8  & GPT-o4-mini       & 35.9 & 25.7 & 71.6 & 9.5 &
19 & Qwen2.5-14B       & 13.0 & 12.7 & 97.3 & 3.7 \\
9  & Claude Sonnet 4   & 26.4 & 23.6 & 89.3 & 7.2 &
20 & Qwen2.5-32B       & 23.2 &  4.2 & 18.2 & 5.5 \\
10 & DeepSeek-V3.2     & 23.9 & 21.8 & 91.2 & 17.0 &
21 & Kimi K2           & 15.1 &  1.4 &  9.3 & 5.3 \\
11 & Gemini 2.0 Flash  & 22.9 & 20.1 & 87.7 & 5.9 &
22 & Llama-3.3-70B     & 11.3 &  0.4 &  3.1 & 2.1 \\
\midrule
\multicolumn{6}{@{}l}{\textit{Average (22 models)}} &
   &                   & 27.6 & 21.3 & 76.3 & 7.8 \\
\bottomrule
\end{tabular}
\end{table}

%% file: tables/sc_trained_results.tex
\begin{table}[t]
\caption{Trained 8B models vs.\ API baselines on \ORSC{} (284 test problems). Domain-specialized 8B models outperform all 22 frontier API models by 38--40 percentage points in \RRR{}, while using 2--3$\times$ fewer steps and 6--19$\times$ fewer tokens.}
\label{tab:sc_trained_results}
\centering
\small
\begin{tabular}{@{}llrrrrrrr@{}}
\toprule
Model & Arch. & P1 \RR{} & \RRR{} & \ME{2} & \ME{4} & \ME{5} & Steps & Tokens \\
\midrule
\Qwen{} (\OptiSTaR{}) & Qwen & 97.2 & \textbf{81.7} & 64.5 & 61.3 & 100.0 & 5.2 & 3,439 \\
\Llama{} (\OptiSTaR{}) & Llama & 97.2 & 80.3 & 67.7 & 74.2 & 100.0 & 4.3 & 3,180 \\
\midrule
Best API (GPT-5.2) & --- & 53.5 & 42.2 & 41.9 & 38.7 & 100.0 & 14.0 & 20,730 \\
Avg.\ API (22 models) & --- & 27.6 & 21.3 & 3.7 & 4.7 & 87.3 & 7.8 & 30,078 \\
\bottomrule
\end{tabular}
\end{table}

%% file: tables/sc_ablation_qwen.tex
\begin{table}[t]
\caption{Component ablation on \ORSC{} (284 test problems). Phase~1 repair is the critical bottleneck: replacing the trained P1 model with API alternatives reduces \RRR{} by 22--40 percentage points. The pattern holds across both architectures.}
\label{tab:sc_ablation}
\centering
\small
\begin{tabular}{@{}llrr@{}}
\toprule
P1 Model & P2 Model & P1 \RR{} (\%) & \RRR{} (\%) \\
\midrule
\multicolumn{4}{@{}l}{\textit{\Qwen{} pipeline (8B)}} \\
\Qwen{} \OptiSTaR{} & \Qwen{} P2 \OptiSTaR{} & 97.2 & \textbf{81.7} \\
\Qwen{} \OptiSTaR{} & Claude Opus 4.5     & 97.2 & 69.4 \\
\Qwen{} \OptiSTaR{} & GPT-5.2             & 96.8 & 68.7 \\
Claude Opus 4.5  & \Qwen{} P2 \OptiSTaR{} & 55.6 & 45.1 \\
GPT-5.2          & \Qwen{} P2 \OptiSTaR{} & 63.0 & 55.6 \\
\midrule
\multicolumn{4}{@{}l}{\textit{\Llama{} pipeline (8B)}} \\
\Llama{} \OptiSTaR{}     & \Llama{} P2 \OptiSTaR{}   & 97.2 & 80.3 \\
\Llama{} \OptiSTaR{}     & Claude Opus 4.5       & 96.8 & 71.1 \\
\Llama{} \OptiSTaR{}     & GPT-5.2               & 97.2 & 71.1 \\
Claude Opus 4.5      & \Llama{} P2 \OptiSTaR{}   & 51.1 & 40.5 \\
GPT-5.2              & \Llama{} P2 \OptiSTaR{}   & 66.5 & 58.1 \\
\midrule
\multicolumn{4}{@{}l}{\textit{Full API (no training)}} \\
GPT-5.2              & GPT-5.2               & 53.5 & 42.3 \\
Claude Opus 4.5      & Claude Opus 4.5       & 33.5 & 30.3 \\
\bottomrule
\end{tabular}
\end{table}

%% file: sections/conclusion.tex
\section{Discussion and Conclusion}
\label{sec:conclusion}

\OptiRepair{} tests whether AI agents can diagnose and repair supply chain optimization models through a closed-loop framework evaluated on 22~API models and two trained 8B-parameter pipelines.
Domain-specialized 8B models achieve 80--82\% \RRR{}, outperforming the best API model (42.2\%) by 38--40pp across both architectures.
The gap concentrates in Phase~1 repair (API average: 27.6\% \RR{} vs.\ 97.2\% trained), while Phase~2 rationality pass rates remain high (76.3\%).
Precise correctness specification, not model capability, determines Phase~2 effectiveness.

\subsection{Managerial Implications}

\paragraph{The recovery bottleneck is solver interaction.}
API models pass Phase~2 rationality checks 76.3\% of the time---frontier LLMs already reason well about supply chain operations.
Yet those same models restore only 27.6\% of infeasible formulations, because iterating on solver feedback, diagnosing \IIS{} conflicts, and proposing targeted constraint repairs is a different skill from answering questions about inventory theory.
Adding supply chain knowledge to prompts does not help reliably: one model gains 7.4pp, another loses 24.1pp.
For operations teams evaluating AI for model maintenance, the scarcest capability is solver interaction---and it responds to targeted training on a modest compute budget.

\paragraph{Feasibility is not rationality: the AI-human decision gap.}
Restoring feasibility without ensuring operational rationality risks producing solutions that appear valid but encode flawed logic---potentially worse than leaving the model for human review.
Roughly one in four API repairs that restore solver feasibility violate an operational property---inverted cost hierarchies, inventory hoarding at upstream echelons, or demand-amplifying orders---that no experienced analyst would accept.
The gap between AI output and expert judgment is not a capability problem; it is a specification problem.
The rationale behind a good supply chain solution---base-stock adherence, echelon coupling, cost consistency---is knowledge that experienced analysts apply implicitly.
AI systems cannot learn it unless it is explicitly codified.
Codifying that rationale as solver-verifiable checks---five predicates grounded in inventory theory \citep{clarkscarf1960, chenbullwhip2000, leebullwhip1997}---improved results more than training larger models.
For organizations adopting AI in operational planning, formalizing what ``rational'' means in their context is the higher-return investment.

\paragraph{Separating repair from validation creates reusable components.}
Because Phase~1 (feasibility repair) is designed as a domain-agnostic solver-interaction module, extending to a new domain requires only a new Phase~2 oracle encoding that domain's correctness criteria.
Organizations can share one repair model across domains while maintaining domain-specific validation.

\subsection{Limitations and Future Work}

Several limitations qualify these findings.
The benchmark uses synthetic errors with serial topology and deterministic demand; real-world practice involves error combinations, network topologies, stochastic demand, and multi-product settings.
The Rationality Oracle encodes operational criteria specific to multi-echelon inventory; organizations with different priorities would need to customize the checks.
All evaluations use greedy decoding, and alternative inference strategies may yield different model rankings.

Expanding the error classification to mixed-integer programs and nonlinear formulations, incorporating real industrial repair traces, and evaluating multi-agent architectures where specialized agents collaborate with domain experts would each extend the evaluation.

For structured operational repair, domain expertise codified as a correctness oracle matters more than model scale---across all error types, architectures, and configurations tested.

%% file: supplement/sections/appendix_a_lp.tex
\section{Full Multi-Echelon LP Formulation}
\label{sec:app_lp}

This appendix details the complete linear programming formulation for the multi-echelon serial supply chain model used in the \ORSC{} evaluation. The formulation follows the echelon-stock framework of \citet{clarkscarf1960} and its LP relaxation. Let $\mathcal{N} = \{1, \ldots, N\}$ denote echelons (1 = retailer, $N$ = factory; $N \in \{2,3,4,5\}$) and $\mathcal{T} = \{1, \ldots, T\}$ denote planning periods ($T \in \{12,16,20,24\}$).

\subsection{Parameters}

\begin{itemize}
\item $h_n \geq 0$: holding cost per unit per period at echelon~$n$. Range: $h_n \in [1, 10]$.
\item $b_n \geq 0$: backorder cost per unit per period at echelon~$n$. Range: $b_n \in [5, 50]$.
\item $C_n > 0$: production or order capacity at echelon~$n$. Range: $C_n \in [50, 500]$.
\item $L_n \geq 0$: replenishment lead time (in periods) at echelon~$n$. Range: $L_n \in \{1, 2, 3\}$.
\item $d_t > 0$: external customer demand at the retailer in period~$t$.
\item $I_{n,0}^{\text{init}} \geq 0$: initial on-hand inventory at echelon~$n$.
\end{itemize}

Demand patterns include three types: (i) stationary ($d_t = \bar{d}$ for all $t$), (ii) step-change ($d_t$ shifts from $\bar{d}_1$ to $\bar{d}_2$ at a change point), and (iii) seasonal ($d_t = \bar{d} + A \sin(2\pi t / T)$ for amplitude $A$).

\subsection{Decision Variables}

\begin{itemize}
\item $x_{n,t} \geq 0$: order quantity placed by echelon~$n$ in period~$t$.
\item $I_{n,t} \geq 0$: on-hand inventory at echelon~$n$ at the end of period~$t$.
\item $B_{n,t} \geq 0$: backorders at echelon~$n$ at the end of period~$t$.
\end{itemize}

\subsection{Objective Function}

The LP minimizes total holding and backorder costs across all echelons and periods:
\begin{equation}
\minimize \quad \sum_{n \in \mathcal{N}} \sum_{t \in \mathcal{T}} \bigl( h_n \, I_{n,t} + b_n \, B_{n,t} \bigr).
\label{eq:app_obj}
\end{equation}

\subsection{Constraints}

\paragraph{C1: Inventory balance (retail echelon).}
For the retailer ($n = 1$):
\begin{equation}
I_{1,t} - B_{1,t} = I_{1,t-1} - B_{1,t-1} + x_{1,t-L_1} - d_t, \quad \forall\, t \in \mathcal{T},
\label{eq:app_inv_retail}
\end{equation}
where $x_{1,t-L_1}$ represents the order placed $L_1$ periods ago that arrives in period~$t$. For $t - L_1 \leq 0$, we set $x_{1,t-L_1} = 0$ (no arrivals from pre-horizon orders unless initial pipeline inventory is modeled separately).

\paragraph{C2: Inventory balance (upstream echelons).}
For each non-retail echelon $n \in \{2, \ldots, N\}$:
\begin{equation}
I_{n,t} - B_{n,t} = I_{n,t-1} - B_{n,t-1} + x_{n,t-L_n} - \sum_{c \in \text{children}(n)} x_{c,t}, \quad \forall\, t \in \mathcal{T},
\label{eq:app_inv_upstream}
\end{equation}
where $\text{children}(n) = \{n{-}1\}$ in the serial topology. The demand faced by echelon~$n$ is the order placed by its downstream neighbor, inducing the propagation mechanism that generates the bullwhip effect \citep{leebullwhip1997, chenbullwhip2000}.

\paragraph{C3: Capacity constraints.}
\begin{equation}
x_{n,t} \leq C_n, \quad \forall\, n \in \mathcal{N},\; t \in \mathcal{T}.
\label{eq:app_capacity}
\end{equation}

\paragraph{C4: Non-negativity.}
\begin{equation}
x_{n,t},\; I_{n,t},\; B_{n,t} \geq 0, \quad \forall\, n \in \mathcal{N},\; t \in \mathcal{T}.
\label{eq:app_nonneg}
\end{equation}

\paragraph{C5: Initial conditions.}
\begin{equation}
I_{n,0} = I_{n,0}^{\text{init}}, \quad B_{n,0} = 0, \quad \forall\, n \in \mathcal{N}.
\label{eq:app_init}
\end{equation}

\paragraph{C6: Demand satisfaction (implicit).}
Backorders $B_{1,t}$ at the retail echelon capture unmet demand. The inventory balance constraint~\eqref{eq:app_inv_retail} ensures that $B_{1,t} \geq d_t - I_{1,t-1} - x_{1,t-L_1} + B_{1,t-1}$ whenever $I_{1,t} = 0$ (i.e., stock is depleted). The backorder cost $b_1$ in the objective penalizes unmet demand, so the LP trades off between holding excess inventory and incurring backorder penalties.

%% file: supplement/sections/appendix_b_errors.tex
\section{Error Type Specifications}
\label{sec:app_errors}

This appendix defines the injection mechanism and solver signature for each of the ten error types introduced in \secref{taxonomy}.
While the main text groups errors by the operational subsystem they affect, \tabref{app_error_specs} documents how each error is injected into a valid LP and how it manifests in Gurobi's \IIS{} output.
The difficulty column reflects empirical solve rates: ``Hard'' errors (\ME{1}, \ME{2}, \ME{4}, \ME{10}) generate misleading \IIS{} signals that require multi-echelon causal reasoning; ``Medium'' errors produce localized but ambiguous signals; ``Easy'' errors (\ME{3}, \ME{5}) yield \IIS{} patterns that directly pinpoint the violated constraint or produce feasible-but-irrational solutions detected by the oracle.

\begin{table}[htbp]
\caption{Error type injection specifications.}
\label{tab:app_error_specs}
\centering
\footnotesize
\begin{tabular}{@{}lp{4.5cm}p{3.5cm}l@{}}
\toprule
Type & Injection Mechanism & IIS Signature & Difficulty \\
\midrule
\ME{1} & Add systematic offset ($3$--$6\times$ demand mean) to demand propagation at one echelon & Multiple \texttt{demand\_prop} constraints + capacity bounds & Hard \\
\ME{2} & Remove arrival term from all inventory balance constraints at target echelon & Multiple \texttt{inv\_balance} constraints + backorder caps & Hard \\
\ME{3} & Force net inventory ($I_{n,t} - B_{n,t}$) to value below achievable minimum & Single \texttt{inv\_balance} constraint with impossible target & Easy \\
\ME{4} & Set order capacity to $0.02$--$0.1\times$ demand mean at retailer & Multiple \texttt{capacity} constraints + inventory balance & Hard \\
\ME{5} & Inflate upstream holding cost to $1.5$--$3.0\times$ downstream cost & None (\OPTIMAL{} but irrational); detected by rationality oracle & Easy \\
\ME{6} & Add constraints forcing $x_{n,t} \geq x_{n-1,t-1} + \text{offset}$ exceeding capacity & Multiple \texttt{bullwhip\_force} constraints + supply cap & Medium \\
\ME{7} & Perturb arrival coefficient from 1.0 to $0.05$--$0.2$ in inventory balance & Multiple \texttt{inv\_balance} constraints with modified coefficient & Medium \\
\ME{8} & Flip sign in demand propagation ($d_{n,t} = -x_{n-1,t}$ instead of $+$) & Multiple \texttt{demand\_prop} constraints (negated term) & Medium \\
\ME{9} & Add minimum-order constraints at factory exceeding supply capacity & Multiple \texttt{min\_order} constraints + supply cap & Medium \\
\ME{10} & Shift time index in demand propagation + add offset ($2$--$4\times$ demand mean) & Multiple \texttt{demand\_prop} constraints (shifted indices) & Hard \\
\bottomrule
\end{tabular}
\end{table}

%% file: supplement/sections/appendix_c_oracle.tex
\section{Phase~II Rationality Checks: Definitions and Thresholds}
\label{sec:app_oracle}

The Phase~II Rationality Oracle evaluates whether a feasible LP solution is operationally rational. It consists of five domain-specific checks, each derived from supply chain theory. Not all checks apply to every error type: the oracle uses an error-type-aware mapping to avoid penalizing correct repairs for artifacts introduced by the error injection process.

\subsection{Five Rationality Checks}

\paragraph{Check 1: Base-stock structure.}
For each echelon~$n$, the oracle verifies that inventory levels approximate the echelon base-stock policy established by \citet{clarkscarf1960}. The check computes the coefficient of variation of inventory:
\begin{equation}
\text{CV}_n = \frac{\text{Std}(I_{n,t})}{\text{Mean}(I_{n,t})},
\label{eq:app_basestock}
\end{equation}
and flags a violation when $\text{CV}_n > \tau_{\text{bs}}$ with $\tau_{\text{bs}} = 2.0$. A well-managed inventory under a base-stock policy exhibits relatively stable levels (low CV), while a poorly repaired model may produce wild inventory fluctuations.

\paragraph{Check 2: Bullwhip ratio.}
The bullwhip effect \citep{leebullwhip1997} measures order variance amplification. For each upstream echelon~$n \geq 2$, the oracle computes:
\begin{equation}
\BWR_n = \frac{\text{Var}(x_{n,t})}{\text{Var}(d_t)},
\label{eq:app_bullwhip}
\end{equation}
where $d_t$ is the external customer demand. The check flags a violation when $\BWR_n > \tau_{\text{bw}}$ with $\tau_{\text{bw}} = 3.0$. The threshold is set above the theoretical lower bound of \citet{chenbullwhip2000} (which depends on lead time and forecasting method) to accommodate the LP's deterministic setting.

\paragraph{Check 3: Inventory allocation.}
In a well-structured serial system, inventory should not concentrate disproportionately at the retailer while upstream echelons hold nothing.
The oracle flags a violation when the retailer's average on-hand inventory exceeds all upstream echelons combined and is meaningful relative to demand:
\begin{equation}
\bar{I}_1 > \tau_{\text{ia}} \cdot \bar{d} \;\;\text{and}\;\; \sum_{n=2}^{N} \bar{I}_n \approx 0,
\label{eq:app_inv_alloc}
\end{equation}
where $\bar{I}_n = \frac{1}{T}\sum_t I_{n,t}$ and $\tau_{\text{ia}} = 0.25$. This check detects structural cost misallocation---particularly cost-swap errors (\ME{5}) where the optimizer places inventory at the wrong echelon because holding cost coefficients are distorted.

\paragraph{Check 4: Cost consistency.}
The oracle enforces two properties of the cost structure.
First, holding costs in the problem configuration should decrease upstream: $h_1 \geq h_2 \geq \cdots \geq h_N$ (with a 1\% tolerance).
Second, if the Gurobi model is available, the oracle checks that each variable's objective coefficient matches the configuration---that is, the model's holding and backorder cost coefficients equal the values specified in the problem parameters.
This check detects repairs that corrupt cost coefficients or tamper with the objective function.

\paragraph{Check 5: Order smoothing.}
The oracle checks that order quantities do not exhibit extreme oscillation:
\begin{equation}
\frac{\max_{t} |x_{n,t} - x_{n,t-1}|}{\text{Mean}(x_{n,t})} < \tau_{\text{os}},
\label{eq:app_order_smooth}
\end{equation}
with $\tau_{\text{os}} = 5.0$. Extreme order oscillation (the ``nervous system'' problem in MRP systems) indicates an unstable replenishment policy that would be operationally infeasible in practice due to production changeover costs and capacity constraints.

\subsection{Which Checks Apply to Which Error Types}

Table~\ref{tab:app_oracle_mapping} specifies which checks apply to each error type. Excluded checks are those where the saboteur's error injection or the associated constraint tightening creates structural conditions that cause the check to fail even on correct repairs.

\begin{table}[htbp]
\caption{Oracle check applicability by error type. A check mark (\checkmark) indicates the check is applied; a dash (--) indicates exclusion. The ``Exclusion rationale'' column explains why excluded checks produce false failures for that error type.}
\label{tab:app_oracle_mapping}
\centering
\small
\begin{tabular}{@{}l ccccc l@{}}
\toprule
Error & \rotatebox{60}{Base-stock} & \rotatebox{60}{Bullwhip} & \rotatebox{60}{Inv.\ alloc.} & \rotatebox{60}{Cost consist.} & \rotatebox{60}{Order smooth.} & Exclusion rationale \\
\midrule
\ME{1} & \checkmark & -- & \checkmark & \checkmark & -- & Demand inflation creates legitimate order variance \\
\ME{2} & \checkmark & -- & \checkmark & \checkmark & -- & Lead-time repair creates legitimate order variance \\
\ME{3} & \checkmark & -- & \checkmark & \checkmark & -- & Balance repair creates legitimate order variance \\
\ME{4} & \checkmark & -- & \checkmark & \checkmark & -- & Capacity change creates legitimate order variance \\
\ME{5} & -- & -- & -- & \checkmark & -- & Cost-only error; cost consistency detects the root cause \\
\ME{6} & -- & -- & -- & -- & \checkmark & Bullwhip error; only order smoothing is artifact-free \\
\ME{7} & -- & -- & -- & \checkmark & -- & Coefficient error; cost consistency detects tampering \\
\ME{8} & -- & -- & -- & \checkmark & -- & Sign error; cost consistency detects coefficient changes \\
\ME{9} & \checkmark & -- & \checkmark & \checkmark & -- & Constraint removal creates legitimate order variance \\
\ME{10} & \checkmark & -- & \checkmark & \checkmark & -- & Index correction creates legitimate order variance \\
\bottomrule
\end{tabular}
\end{table}

%% file: supplement/sections/appendix_k_prompts.tex
\section{Prompt and Interaction Examples}
\label{sec:app_prompts}

This appendix presents the full system prompt used by \OptiRepair{} agents and a complete interaction transcript illustrating both phases of the closed-loop pipeline.

\subsection{System Prompt}
\label{sec:app_system_prompt}

The following system prompt is used for trained models (\Qwen{} and \Llama{}). API models receive a JSON-formatted variant with identical content.

\begin{systempromptbox}
You are an expert supply chain optimization debugger.
Your task is to debug infeasible multi-echelon supply
chain LP models through a two-phase closed-loop process.

\medskip
\textbf{== Two-Phase Process ==}

\textbf{Phase 1 (Debug):} Fix infeasibility. Identify
conflicting constraints and repair them to restore
feasibility.

\textbf{Phase 2 (Validate):} After restoring feasibility,
an automated rationality oracle checks the solution
for operational rationality. A solver-feasible solution
is not necessarily a good supply chain policy. Checks:
1. Base-stock rationality -- cost-efficient inventory\\
2. Bullwhip control -- order variance within bounds\\
3. Monotonicity -- no severe inventory inversions\\
4. Cost consistency -- coefficients match specification\\
5. Order smoothing -- no excessive order oscillation

If the solution fails, you receive violation feedback
and must fix the structural issue.

\medskip
\textbf{== Available Actions ==}
- GET\_IIS(): Compute Irreducible Infeasible Subsystem\\
- DROP\_CONSTRAINT(name): Remove a constraint\\
- RELAX\_CONSTRAINT(name, amount): Relax a constraint\\
- UPDATE\_RHS(name, value): Update right-hand side\\
- UPDATE\_OBJ(var\_prefix, value): Update objective coeff\\
- UPDATE\_BOUNDS(var, lb, ub): Update variable bounds\\
- CHECK\_SLACK(name): Check slack for a constraint\\
- SUBMIT(): Submit the repaired model

\medskip
\textbf{== Response Format ==}

<think>\\
Your reasoning about the infeasibility or\\
rationality violation...\\
</think>

Action: ACTION\_TYPE(arguments)
\end{systempromptbox}

\subsection{State Observation Format}
\label{sec:app_state_format}

At each step, the agent receives a structured observation containing the solver status, \IIS{} information (if applicable), model dimensions, and recent action history.

\begin{envbox}[State Observation]
\#\# Current State\\
- Solver Status: INFEASIBLE\\
- Step: 0

\medskip
\#\# IIS (Irreducible Infeasible Subsystem)\\
- Conflicting Constraints: ['capacity\_e1\_t1',\\
\hspace*{1em}'capacity\_e1\_t3', 'capacity\_e1\_t5',\\
\hspace*{1em}'demand\_balance\_e1\_t1', 'demand\_balance\_e1\_t3']\\
- Conflicting Bounds: ['x\_e1\_t1 <= 20',\\
\hspace*{1em}'x\_e1\_t3 <= 20', 'x\_e1\_t5 <= 20']

\medskip
\#\# Model Structure\\
- Total Constraints: 142\\
- Total Variables: 96\\
- Constraint Names (first 10): ['demand\_balance\_e1\_t1',\\
\hspace*{1em}'demand\_balance\_e1\_t2', ..., 'capacity\_e1\_t1', ...]

\medskip
What action should be taken next?
\end{envbox}

\subsection{Phase~I Interaction Transcript}
\label{sec:app_phase1_transcript}

The following transcript shows the trained \Qwen{} model debugging an \ME{4} error (capacity too tight) on a 3-echelon, 16-period supply chain.

\begin{agentbox}[Agent -- Turn 1]
<think>\\
The model is infeasible. The IIS shows capacity
constraints at echelon 1 (capacity\_e1\_t1, t3, t5)
conflicting with demand balance constraints. The
variable bounds confirm x\_e1 <= 20 for all flagged
periods. Average demand is 45 units, well above the
capacity of 20. This is an ME-4 error: capacity set
far below demand, causing cascading infeasibility.
I should relax the echelon-1 capacity constraints.\\
</think>

\medskip
Action: RELAX\_CONSTRAINT(capacity\_e1, 50.0)
\end{agentbox}

\begin{envbox}[Environment Response]
\#\# Current State\\
- Solver Status: OPTIMAL\\
- Step: 1\\
- Objective Value: 12487.3

\medskip
-> Entering Phase 2 (Rationality Oracle)
\end{envbox}

\subsection{Phase~II Interaction Transcript}
\label{sec:app_phase2_transcript}

After Phase~I restores feasibility, the rationality oracle evaluates the solution.

\begin{oraclebox}[Rationality Oracle -- Feedback]
\#\# Rationality Feedback (from Phase 2 Oracle)

Echelon 1 (Retailer, h=\$8.00) holds 45.2 units
on-hand inventory while Echelon 2 (Warehouse,
h=\$3.00) holds 12.3 units. Holding cost at Retailer
(\$8.00) > Warehouse (\$3.00). Cost-efficient
allocation violated.

\medskip
The solver found a feasible solution, but it failed
rationality checks. The repair restored feasibility
but created an operationally irrational policy.
Fix the structural issue described above, then
restore feasibility again.

\medskip
\#\# Closed-Loop Status\\
- Phase: VALIDATE -> DEBUG (loop-back)\\
- Debug-Validate Loop: 1
\end{oraclebox}

\begin{agentbox}[Agent -- Turn 2]
<think>\\
The capacity relaxation fixed infeasibility but
broke cost consistency. The retailer (echelon 1)
has holding cost h=\$8 but holds 45 units, while
the warehouse (echelon 2) with h=\$3 holds only 12
units. In a rational supply chain, the cheaper
storage location should buffer more inventory.
The holding cost coefficient for echelon 1 in the
objective function may be distorted. I should
update it to match the warehouse's cost structure.\\
</think>

\medskip
Action: UPDATE\_OBJ(hold\_e1, 3.0)
\end{agentbox}

\begin{envbox}[Environment Response]
\#\# Current State\\
- Solver Status: OPTIMAL\\
- Step: 2\\
- Objective Value: 11832.1

\medskip
-> Re-checking Rationality Oracle
\end{envbox}

\begin{oraclebox}[Rationality Oracle -- Result]
\#\# Rationality Check: PASSED\\
- Base-stock rationality: PASS\\
- Bullwhip control: PASS\\
- Monotonicity: PASS\\
- Cost consistency: PASS\\
- Order smoothing: PASS

\medskip
Episode complete. Solution is both solver-feasible
and operationally rational.\\
Total steps: 2 | Loops: 1
\end{oraclebox}

\subsection{API Model Response Format}
\label{sec:app_api_format}

API models (GPT-5.2, Claude, Gemini, etc.) use a JSON response format instead of the \texttt{<think>}/\texttt{Action:} format:

\begin{agentbox}[API Agent Response (JSON)]
\{\\
\hspace*{1em}"reasoning": "The IIS shows capacity constraints\\
\hspace*{3em}at echelon 1 conflicting with demand balance.\\
\hspace*{3em}Capacity C1=20 is below demand of 45. I should\\
\hspace*{3em}relax the capacity constraint.",\\
\hspace*{1em}"action": "RELAX\_CONSTRAINT",\\
\hspace*{1em}"target": "capacity\_e1",\\
\hspace*{1em}"value": 50.0\\
\}
\end{agentbox}

\noindent Both formats are parsed into identical \texttt{Action} objects by the environment's action parser.
The trained model format is more token-efficient (no JSON overhead) and allows natural chain-of-thought reasoning within the \texttt{<think>} tags.

%% file: supplement/sections/appendix_e_full_results.tex
\section{Full Per-Error-Type Results}
\label{sec:app_full_results}

This appendix reports the complete per-error-type breakdown for all models evaluated on \ORSC{}.
\figref{p1_vs_p2} in the main text decomposes overall performance into Phase~1 Recovery Rate and Phase~2 Pass Rate; the tables below give the per-error-type detail.
\tabref{sc_per_me} presents per-error-type \RRR{} for both trained models and the top-6 API baselines.
\tabref{app_full_per_me} presents the complete \RRR{} breakdown for all 22 API models in landscape format.

\input{tables/sc_per_me}

\begin{landscape}
\begin{table}[!ht]
\caption{\RRR{} (\%) by error type for all 22 API models and the trained \Qwen{} pipeline on \ORSC{} test set. Bold = highest API score per error type.}
\label{tab:app_full_per_me}
\centering
\footnotesize
\begin{tabular}{@{}rl rrrrr rrrrr r@{}}
\toprule
Rank & Model & \ME{1} & \ME{2} & \ME{3} & \ME{4} & \ME{5} & \ME{6} & \ME{7} & \ME{8} & \ME{9} & \ME{10} & Overall \\
\midrule
1 & GPT-5.2 & 7.4 & \textbf{41.9} & 73.3 & \textbf{38.7} & 100 & 7.1 & 50.0 & 36.0 & \textbf{57.1} & 6.7 & 42.2 \\
2 & Gemini 2.5 Pro & 0.0 & 6.5 & 63.3 & 6.5 & 100 & \textbf{60.7} & 50.0 & \textbf{60.0} & \textbf{57.1} & 23.3 & 42.2 \\
3 & DeepSeek-R1-0528 & 3.7 & 12.9 & 76.7 & 9.7 & 100 & 0.0 & \textbf{58.3} & \textbf{60.0} & 42.9 & 10.0 & 37.0 \\
4 & Gemini 2.5 Flash & 0.0 & 3.2 & 66.7 & 9.7 & 100 & 21.4 & 41.7 & 44.0 & 21.4 & 13.3 & 32.0 \\
5 & Qwen2.5-7B & \textbf{40.7} & 0.0 & 66.7 & 9.7 & 90.0 & 3.6 & 29.2 & 36.0 & 0.0 & \textbf{33.3} & 31.0 \\
6 & Claude Opus 4.5 & 3.7 & 6.5 & 70.0 & 3.2 & 100 & 35.7 & 37.5 & 20.0 & 25.0 & 0.0 & 30.3 \\
7 & Qwen2.5-Max & 18.5 & 0.0 & 63.3 & 9.7 & 100 & 0.0 & 50.0 & 16.0 & 0.0 & 10.0 & 26.8 \\
8 & GPT-o4-mini & 7.4 & 0.0 & 50.0 & 0.0 & 100 & 7.1 & 25.0 & 16.0 & 39.3 & 10.0 & 25.7 \\
9 & Claude Sonnet 4 & 3.7 & 6.5 & 66.7 & 3.2 & 100 & 0.0 & 12.5 & 8.0 & 25.0 & 3.3 & 23.6 \\
10 & DeepSeek-V3.2 & 3.7 & 0.0 & 50.0 & 3.2 & 100 & 10.7 & 12.5 & 20.0 & 14.3 & 0.0 & 21.8 \\
11 & Gemini 2.0 Flash & 0.0 & 0.0 & \textbf{80.0} & 0.0 & 100 & 0.0 & 8.3 & 0.0 & 3.6 & 0.0 & 20.1 \\
12 & Claude Haiku 4.5 & 0.0 & 0.0 & 23.3 & 0.0 & 100 & 7.1 & 25.0 & 28.0 & 7.1 & 0.0 & 19.0 \\
13 & GPT-o3 & 0.0 & 0.0 & 36.7 & 3.2 & 100 & 0.0 & 8.3 & 0.0 & 32.1 & 3.3 & 19.0 \\
14 & GPT-4.1 & 0.0 & 3.2 & 36.7 & 6.5 & 100 & 0.0 & 8.3 & 4.0 & 10.7 & 0.0 & 17.6 \\
15 & GPT-4.1 Mini & 3.7 & 0.0 & 26.7 & 0.0 & 100 & 0.0 & 4.2 & 0.0 & 32.1 & 0.0 & 17.2 \\
16 & GPT-5 Mini & 0.0 & 0.0 & 33.3 & 0.0 & 100 & 0.0 & 12.5 & 0.0 & 14.3 & 3.3 & 16.9 \\
17 & GPT-o1 & 0.0 & 0.0 & 20.0 & 0.0 & 96.7 & 0.0 & 4.2 & 0.0 & 7.1 & 0.0 & 13.5 \\
18 & Claude 3.7 Sonnet & 0.0 & 0.0 & 16.7 & 0.0 & 100 & 0.0 & 8.3 & 0.0 & 3.6 & 0.0 & 13.4 \\
19 & Qwen2.5-14B & 0.0 & 0.0 & 16.7 & 0.0 & 100 & 0.0 & 4.2 & 0.0 & 0.0 & 0.0 & 12.7 \\
20 & Qwen2.5-32B & 0.0 & 0.0 & 16.7 & 0.0 & 20.0 & 0.0 & 0.0 & 0.0 & 0.0 & 3.3 & 4.2 \\
21 & Kimi K2 & 0.0 & 0.0 & 3.3 & 0.0 & 10.0 & 0.0 & 0.0 & 0.0 & 0.0 & 0.0 & 1.4 \\
22 & Llama-3.3-70B & 0.0 & 0.0 & 0.0 & 0.0 & 3.3 & 0.0 & 0.0 & 0.0 & 0.0 & 0.0 & 0.4 \\
\midrule
--- & Trained \Qwen{} & 85.2 & 64.5 & 96.7 & 61.3 & 100 & 89.3 & 91.7 & 88.0 & 96.4 & 86.7 & 81.7 \\
\bottomrule
\end{tabular}
\end{table}
\end{landscape}

%% file: tables/sc_per_me.tex
\begin{table}[!htb]
\caption{Per-error-type \RRR{} (\%) on \ORSC{} for trained models and top-6 API baselines. \ME{5} (cost structure) is universally solvable; \ME{2} (lead time) and \ME{4} (capacity) separate trained models from API baselines.}
\label{tab:sc_per_me}
\centering
\footnotesize
\begin{tabular}{@{}lrrrrrrrrrr@{}}
\toprule
Model & \ME{1} & \ME{2} & \ME{3} & \ME{4} & \ME{5} & \ME{6} & \ME{7} & \ME{8} & \ME{9} & \ME{10} \\
\midrule
\Qwen{} (trained)       & 85.2 & 64.5          & 96.7                  & 61.3           & \textbf{100} & \textbf{89.3} & \textbf{91.7} & \textbf{88.0} & \textbf{96.4} & \textbf{86.7} \\
\Llama{} (trained)      & \textbf{81.5} & \textbf{67.7} & 93.3         & \textbf{74.2}  & \textbf{100} & 85.7          & 87.5          & 84.0          & 92.9          & 83.3 \\
\midrule
GPT-5.2                 &  7.4 & 41.9          & 73.3 & 38.7           & \textbf{100} &  7.1          & 50.0          & 36.0          & 57.1 &  6.7 \\
Gemini 2.5 Pro          &  0.0 &  6.5          & 63.3 &  6.5           & \textbf{100} & 60.7 & 50.0          & 60.0 & 57.1 & 23.3 \\
DeepSeek-R1             &  3.7 & 12.9          & 76.7 &  9.7           & \textbf{100} &  0.0          & 58.3 & 60.0 & 42.9          & 10.0 \\
Gemini 2.5 Flash        &  0.0 &  3.2          & 66.7 &  9.7           & \textbf{100} & 21.4          & 41.7          & 44.0          & 21.4          & 13.3 \\
Claude Opus 4.5         &  3.7 &  6.5          & 70.0 &  3.2           & \textbf{100} & 35.7          & 37.5          & 20.0          & 25.0          &  0.0 \\
Qwen2.5-7B              & 40.7 & 0.0           & 66.7 &  9.7           &  90.0        &  3.6          & 29.2          & 36.0          &  0.0          & 33.3 \\
\bottomrule
\end{tabular}
\end{table}

%% file: supplement/sections/appendix_f_guided.tex
\section{Guided Prompt Experiments}
\label{sec:app_guided}

We evaluated two guided prompt variants on Claude Opus~4.5 and GPT-5.2-chat. The \emph{anti-loop} variant added explicit instructions to avoid repeating failed repair strategies; the \emph{domain-knowledge} variant injected supply chain structural hints (base-stock properties, bullwhip ratio interpretation, cost structure constraints). The domain-knowledge prompt improved Opus by 7.4 percentage points (\RRR{} 30.3\% to 37.7\%) but degraded GPT-5.2 by 24.1 percentage points (42.2\% to 18.1\%). The anti-loop prompt hurt both models. Prompt text is available from the authors upon request.

%% file: supplement/sections/appendix_i_dataset.tex
\section{Dataset Construction Details}
\label{sec:app_dataset}

This appendix describes the generation pipeline, instance parameters, validation procedures, and dataset statistics for \ORSC{}.

\subsection{\ORSC{} Generation Pipeline}

We construct the \ORSC{} problem set through a four-stage pipeline:

\begin{enumerate}
\item \textbf{Source instance generation.} We generate supply chain LP instances by sampling parameters from the ranges specified in Table~\ref{tab:app_instance_params}. Each instance is solved to optimality by Gurobi 12.0 to confirm a feasible baseline. Trivially feasible instances (optimal objective value below a threshold) or degenerate ones (fewer than 10 active constraints at optimality) are filtered out.

\item \textbf{Sabotage.} For each valid source instance, the saboteur selects an error type (\ME{1}--\ME{10}) and applies the corresponding injection mechanism (Appendix~\ref{sec:app_errors}). The saboteur first tightens the model with operational bounds---backorder caps and factory supply caps calibrated to the optimal solution---to prevent the LP's natural slack from absorbing the perturbation. The error is then injected into the tightened model.

\item \textbf{Gurobi verification.} Each sabotaged model is solved by Gurobi. The model must return \INFEASIBLE{} status (or \OPTIMAL{}-but-irrational for \ME{5}), and the \IIS{} computation must return a non-empty subsystem. Models that fail either check are discarded.

\item \textbf{Oracle labeling.} The ground-truth fix is applied, and the repaired model is re-solved. The solution must achieve \OPTIMAL{} status and pass all applicable Rationality Oracle checks (Appendix~\ref{sec:app_oracle}). Problems where the ground-truth fix does not pass all checks are discarded.
\end{enumerate}

\subsection{Instance Parameters}

\begin{table}[htbp]
\caption{Parameter ranges for \ORSC{} instance generation.}
\label{tab:app_instance_params}
\centering
\small
\begin{tabular}{@{}llr@{}}
\toprule
Parameter & Range & Distribution \\
\midrule
Number of echelons $N$ & $\{2, 3, 4, 5\}$ & Uniform \\
Planning periods $T$ & $\{12, 16, 20, 24\}$ & Uniform \\
Topology & Serial & Fixed \\
Holding cost $h_n$ & $[1, 10]$ per unit/period & Uniform \\
Backorder cost $b_n$ & $[5, 50]$ per unit/period & Uniform \\
$h$-to-$b$ ratio & $1{:}2$ to $1{:}10$ & Derived \\
Capacity $C_n$ & $[50, 500]$ units/period & Uniform \\
Lead time $L_n$ & $\{1, 2, 3\}$ periods & Uniform \\
Initial inventory $I_{n,0}^{\text{init}}$ & $[0, 2 \bar{d}]$ & Uniform \\
Mean demand $\bar{d}$ & $[50, 200]$ units/period & Uniform \\
Demand pattern & Stationary, step-change, seasonal & Categorical (equal) \\
\bottomrule
\end{tabular}
\end{table}

\subsection{Dataset Statistics}

\begin{table}[htbp]
\caption{Dataset split statistics for \ORSC{}.}
\label{tab:app_dataset_stats}
\centering
\small
\begin{tabular}{@{}l rr r@{}}
\toprule
& Train & Test & Total \\
\midrule
Total problems & 692 & 284 & 976 \\
\midrule
\ME{1} (Demand Inflation) & 78 & 27 & 105 \\
\ME{2} (Lead Time Error) & 89 & 31 & 120 \\
\ME{3} (Balance Violation) & 87 & 30 & 117 \\
\ME{4} (Capacity Reduction) & 89 & 31 & 120 \\
\ME{5} (Cost Structure Error) & 71 & 30 & 101 \\
\ME{6} (Bullwhip Amplification) & 40 & 28 & 68 \\
\ME{7} (Coefficient Perturbation) & 66 & 24 & 90 \\
\ME{8} (Sign Error) & 71 & 25 & 96 \\
\ME{9} (Redundant Constraint) & 56 & 28 & 84 \\
\ME{10} (Index Mismatch) & 45 & 30 & 75 \\
\midrule
Avg.\ variables per instance & \multicolumn{2}{c}{185} & \\
Avg.\ constraints per instance & \multicolumn{2}{c}{162} & \\
Avg.\ IIS size & \multicolumn{2}{c}{12.3 constraints} & \\
\bottomrule
\end{tabular}
\end{table}

The train/test split is stratified by error type, with disjoint supply chain instances (i.e., no source instance appears in both splits). The test set contains 284 problems, providing sufficient statistical power to detect differences of 5 percentage points or larger at 95\% confidence.